%% file: sample-acmsmall.tex
\documentclass[acmsmall]{acmart}

\AtBeginDocument{%
  \providecommand\BibTeX{{%
    \normalfont B\kern-0.5em{\scshape i\kern-0.25em b}\kern-0.8em\TeX}}}

\setcopyright{acmlicensed}
\acmJournal{TOMM}
\acmYear{2021} \acmVolume{1} \acmNumber{1} \acmArticle{1} \acmMonth{1} \acmPrice{15.00}\acmDOI{10.1145/3472768}

\acmJournal{TOMM}
\acmVolume{37}
\acmNumber{4}
\acmArticle{111}
\acmMonth{8}

\usepackage[utf8]{inputenc} 
\usepackage[T1]{fontenc}    

\usepackage{microtype}
\usepackage{graphicx}
\usepackage{url}            
\usepackage{amsfonts}       
\usepackage{nicefrac}       
\usepackage{xcolor}
\usepackage{enumitem}
\usepackage[noend]{algpseudocode}
\usepackage{algorithm}
\usepackage{epsfig}
\usepackage{tabularx}
\usepackage{graphicx}
\usepackage{subfig}
\usepackage{wrapfig}
\usepackage{textcomp}
\usepackage{siunitx}
\usepackage{color}
\usepackage{bigints}
\usepackage{booktabs} 
\usepackage{ulem}
\usepackage{bbm}
\usepackage{amssymb}
\usepackage{multirow}
\usepackage{soul}
\usepackage{wrapfig}
\usepackage{hyperref}
\usepackage{xpatch}
\usepackage{amsmath}

\usepackage[font=small,labelfont=bf]{caption}

\usepackage{fancyhdr}
\usepackage{xspace}

\DeclareMathOperator*{\argmax}{arg\,max}

\let\emptyset\varnothing

\renewcommand{\algorithmiccomment}[1]{\bgroup\hfill\small//~#1\egroup}
\algnewcommand\algorithmicforeach{\textbf{for each}}
\algdef{S}[FOR]{ForEach}[1]{\algorithmicforeach\ #1\ \algorithmicdo}

\makeatletter
\DeclareRobustCommand\onedot{\futurelet\@let@token\@onedot}
\def\@onedot{\ifx\@let@token.\else.\null\fi\xspace}
\def\eg{\textit{e.g}\onedot} \def\Eg{\textit{E.g}\onedot}
\def\ie{\textit{i.e}\onedot} 
 
\def\etc{\textit{etc}\onedot} 
\def\wrt{w.r.t\onedot} 
\def\etal{\textit{et al}\onedot}
\makeatother

\makeatletter
\renewcommand\@formatdoi[1]{\ignorespaces}
\makeatother

\makeatletter
\let\@authorsaddresses\@empty
\makeatother

\begin{document}

\title{Black-Box Diagnosis and Calibration on GAN Intra-Mode Collapse: A Pilot Study}

\author{Zhenyu Wu}
\email{wuzhenyu\_sjtu@tamu.edu}
\affiliation{%
  \institution{Texas A\&M University}
  \streetaddress{400 Bizzell St}
  \city{College Station}
  \state{TX}
  \country{USA}
  \postcode{77843}
}

\author{Zhaowen Wang}
\email{zhawang@adobe.com}
\affiliation{%
  \institution{Adobe Research}
  \streetaddress{345 Park Avenue}
  \city{San Jose}
  \state{CA}
  \country{USA}
  \postcode{95110}
}

\author{Ye Yuan}
\email{ye.yuan@tamu.edu}
\affiliation{%
  \institution{Texas A\&M University}
  \streetaddress{400 Bizzell St}
  \city{College Station}
  \state{TX}
  \country{USA}
  \postcode{77843}
}

\author{Jianming Zhang}
\email{jianmzha@adobe.com}
\affiliation{%
  \institution{Adobe Research}
  \streetaddress{345 Park Avenue}
  \city{San Jose}
  \state{CA}
  \country{USA}
  \postcode{95110}
}

\author{Zhangyang Wang}
\email{atlaswang@utexas.edu}
\affiliation{%
  \institution{The University of Texas at Austin}
  \streetaddress{Main Building (MAI) 110 Inner Campus Drive}
  \city{Austin}
  \state{TX}
  \country{USA}
  \postcode{78705}
}

\author{Hailin Jin}
\email{hljin@adobe.com}
\affiliation{%
  \institution{Adobe Research}
  \streetaddress{345 Park Avenue}
  \city{San Jose}
  \state{CA}
  \country{USA}
  \postcode{95110}
}

\input{chapters/0_Abstract}




\maketitle

\input{chapters/1_Introduction}
\input{chapters/2_RelatedWorks}
\input{chapters/3_Method}
\input{chapters/4_Experiments}
\input{chapters/5_Limitations}

\bibliographystyle{bib_style/ACM-Reference-Format}
\bibliography{sample-acmsmall}

\end{document}

%% file: chapters/0_Abstract.tex
\begin{abstract}
Generative adversarial networks (GANs) nowadays are capable of producing images of incredible realism. 
One concern raised is whether the state-of-the-art GAN's learned distribution still suffers from mode collapse, and what to do if so. Existing diversity tests of samples from GANs are usually conducted qualitatively on a small scale, and/or depends on the access to original training data as well as the trained model parameters. This paper explores to diagnose GAN \textit{intra-mode collapse} and calibrate that, in a novel \textit{black-box} setting: no access to training data, nor the trained model parameters, is assumed. The new setting is practically demanded, yet rarely explored and significantly more challenging. As a first stab, we devise a set of statistical tools based on sampling, that can visualize, quantify, and rectify \textit{intra-mode collapse}. We demonstrate the effectiveness of our proposed diagnosis and calibration techniques, via extensive simulations and experiments, on unconditional GAN image generation (\eg, face and vehicle). Our study reveals that the \textit{intra-mode collapse} is still a prevailing problem in state-of-the-art GANs and the mode collapse is diagnosable and calibratable in \textit{black-box} settings. Our codes are available at: {\emph{\url{https://github.com/VITA-Group/BlackBoxGANCollapse}}}.
\end{abstract}

%% file: chapters/1_Introduction.tex
\section{Introduction}
Generative adversarial networks (GANs)~\citep{goodfellow2014generative,karras2017progressive,karras2018style,karras2019analyzing,arjovsky2017wasserstein,mao2017least,wu2020mm,gong2019autogan,jiang2021enlightengan,kupyn2019deblurgan,jiang2021transgan,wang2020gan,yang2021shape,yang2019controllable,yang2020deep} have demonstrated unprecedented power for image generation. However, GANs suffer from generation bias and/or loss of diversity. The underlying reasons could be compound, ranging from the data imbalance to the training difficulty, and more:
\begin{itemize}
    \item First, the training data for GANs, \eg, for the most-studied \textit{unconditional/unsupervised} generation tasks~\citep{karras2017progressive,karras2018style}, could possess various subject or attribute imbalances~\citep{wang2020revise}. Consequently, GANs trained with them might be further biased towards the denser areas, similarly to the classifier bias towards the majority class in imbalanced classification.
    \item More intrinsically, even when the training data ``looks'' balanced, training GANs is much more unstable and uncontrollable than training classifiers. One common hurdle of GANs is the loss of diversity due to mode collapse~\citep{goodfellow2016nips}, wherein the generator concentrates a probability mass on only a few modes of the true distribution. It is often considered as a \textit{training artifact}. Another widely reported issue, covariate shift~\citep{santurkar2017classification}, could be viewed as a nuanced version of mode collapse.
\end{itemize}

\subsection{Diversity Evaluation of GANs}
There are several popular metrics for evaluation, \eg, Inception Score (IS)~\citep{salimans2016improved}, Fréchet Inception Distance (FID)~\citep{heusel2017gans}, MODE~\citep{che2016mode} and birthday paradox test~\citep{arora2017gans}. However, they are not always sensitive to mode collapses; see Section 2.1 for more discussions.

Recently, two classification-based metrics~\citep{santurkar2017classification,bau2019seeing} have been proposed for quantitatively assessing the mode distribution learned by GANs by comparing the learned distribution of attributes or objects in the generated images with the target distribution in the training set. However, Santurkar~\etal~\citep{santurkar2017classification} hinges on a classifier trained on the original (and manually balanced) GAN training set, with class labels known, available, and well-defined (\eg, object classes in CIFAR-10, or facial attributes in CelebA); Bau~\etal~\citep{bau2019seeing} relies on an instance segmentation model trained on images with $336$ object classes. They mainly focus on detecting the mode collapse measured at class level but do not directly address the mode collapse within a class.

Mode collapse refers to the limited sample variety in the generator's learned distribution. As discussed by Huang~\etal~\citep{huang2018introduction}, \textit{inter-mode collapse} occurs when some modes (\eg, digit classes in MNIST) are never produced from the generated samples; while \textit{intra-mode collapse} occurs when all modes (\eg, classes) can be found in the generated samples but with limited variations (\eg, a digit with few writing styles).
Both types of mode collapses are commonly observed in GANs, and the above classification-based metrics, by definition, can only detect the inter-mode collapse. Also, they cannot be easily extended to images subjects where classes are not well defined and/or not enumerable (\eg, the identity of generated faces/vehicles, or in other open set problems). 

Beyond the above progress, many open questions persist, including but not limited to: \textit{do state-of-the-art GANs still suffer from \textit{intra-mode collapse}?  Can we detect it with minimal assumptions or efforts? Moreover, whether there is an ``easy and quick'' remedy to alleviate it?} -- \underline{Addressing them motivates our work}.

\subsection{Black-Box Diagnosis \& Calibration}
Many approaches have been proposed to alleviate mode collapse, ranging from better optimization objectives~\citep{arjovsky2017wasserstein, mao2017least} to customized building blocks~\citep{durugkar2016generative, ghosh2018multi, liu2016coupled, lin2018pacgan}. However, they require either specialized GAN architectures and/or tedious (re-)training, or at least access to training data and/or model parameters. Whether it is for \textit{diagnosing} (detecting or evaluating) mode collapse, or \textit{calibrating} (alleviating or fixing) it, all the above methods require the original training data and/or the trained model parameters. We refer to those methods as \textbf{white-box} approaches. 

In contrast, the main goal of this work is to significantly extend the applicability of such diagnosis and calibration to an almost unexplored \textbf{black-box} setting: \textit{we assume neither access to the original training data, nor the model parameters, nor the class labels of the original data (which might be inaccessible or even not well defined, as above explained)}. To our best knowledge, no existing approach is immediately available to address this new challenge. Instead, we find such \textit{black-box} setting desired by practitioners due to the following reasons: (i) the training data might be protected or no longer available since it contains sensitive information (\textit{e.g.}, human faces or person images); (ii) the GAN model might be provided as a black box and cannot be modified (\textit{e.g.}, as commercial IPs, executables, or APIs); (iii) the practitioners want to adjust the generated distribution of any GAN without expensive re-training, to enable fast turn-around and also save training resources.

Assume, for just one example, a GAN model is protected by IP and provided to users as an executable (or cloud API) only. The \textit{black-box} diagnosis and calibration are helpful for both the users and the provider. \ul{For the users}, they could effectively discover whether the provided API displays any unexpected generation deficiency or bias, despite having no access to the weights nor data. \ul{For the provider}, they could identify a collapse and quickly fix it, by adding merely light-weight ``wrapping'' (\eg, output post-processing) to the model, instead of costly (even infeasible) re-training.

\subsection{Our Contributions}
As a first stab at this new challenge, we propose hypothesis testing methods to analyze the clustering density pattern of generated samples. To characterize point patterns over a given area of interest, we incorporate statistical tools in spatial analysis and the Monte Carlo method to provide both qualitative and quantitative measures of mode collapse. Then, for the first time, we explore two \textit{black-box} approaches to calibrate the GAN's learned distribution and to rectify the detected mode collapse, via Gaussian mixture models and importance sampling, respectively. It is crucial to notice that neither the proposed diagnosis nor the calibrations touch the original training data, access the model parameters, or re-train the model in any way: they are all based on \textit{sampling from the ``black box''}.

We demonstrate the application of our new toolkit in analyzing the \textit{intra-mode collapse} in \textit{unconditional image generation} tasks, such as face and vehicle images.
Instead of measuring a ``global'' class distribution, our method focuses on addressing ``local'' high-density regions. Therefore, it is specialized at detecting the \textit{intra-mode collapse}, and is complementary to~\citep{santurkar2017classification} and~\citep{bau2019seeing}. 
We find the \textit{intra-mode collapse} remains a prevailing problem in state-of-the-art GANs~\citep{karras2018style,karras2017progressive,brock2018large,karras2019analyzing}. We analyze several possible causes and demonstrate our calibration approaches can notably alleviate the issue.

Although beyond our discussion scope, we point out that our proposed diagnosis and calibration on intra-mode collapse can contribute to understanding the privacy~\citep{filipovych2011semi,zheng2020pcal,wu2018towards,wu2020privacy} and fairness~\citep{holstein2018improving,wu2019delving,uplavikar2019all} issues in generative models. First, the collapsed mode in GAN's learned distribution, \ie, images of repeated identity, could focus on some training data, especially when the data is highly imbalanced, thus causing privacy breach if the training data is protected. Second, the collapsed mode shows the generative model's bias towards some specific identities. Many existing works using generated synthetic images together with or instead of real images for training, with their purposes ranging from semi-supervised learning \cite{salimans2016improved} to small data augmentation \cite{bowles2018gansfer,liu2019pixel,zhang2019dada}. As a potential consequence, training with the generated data might incur biased classifier predictions.

%% file: chapters/2_RelatedWorks.tex
\section{Related Works}
\subsection{Privacy and Fairness Concerns in GANs}
\subsubsection{Privacy}
The privacy breach risk of GANs lies in generating data that are more likely to be substantially similar to existing training samples, as a consequence of potential overfitting. Xie~\etal~\cite{xie2018differentially} argues that the density of the learned distribution could overly concentrate on the training data points, which is alarming when the training data is private or sensitive. The authors proposed to train a differential private GAN (DPGAN) by gradient noise injection and then to clip. Webster~\etal~\cite{webster2019detecting} studies GAN's overfitting issue by analyzing the statistics of reconstruction errors on both training and validation images, by optimizing the latent code to find the nearest neighbor in the generation manifold. Their empirical study finds out that standard GAN evaluation metrics often fail to capture memorization for deep generators, making overfitting undetectable for pure GANs and causing privacy leak risks.

\subsubsection{Fairness}
Amini~\etal~\cite{amini2019uncovering} propose an debiasing VAE
(DB-VAE) algorithm, on mitigating generation bias, but it needs a large dataset to learn its latent structure. Xu~\etal~\cite{xu2018fairgan} develops FairGAN for fair data generation, which achieves statistical parity with regard to a protected attribute, using an auxiliary discriminator to ensure no correlation between protected/unprotected attributes, as well as between the utility task and the protected attribute. Sattigeri~\etal~\cite{sattigeri2018fairness} also aimed to generate debiased, fair data to protected attributes in allocative decision making, with a pair of auxiliary losses introduced to encourage demographic parity. Unlike those existing works, we seek to analyze and gain insights into the fairness issue in current state-of-the-art GANs (rather than specifically crafted ones), where currently no fairness constraint has not been, or is non-trivial to be enforced.


\subsection{Evaluation of Mode Collapse in GANs}
GANs are often observed to suffer from the mode collapse problem~\citep{salimans2016improved,sutskever2015towards}, where only a small subset of distribution modes are characterized by the generator. The problem is especially prevalent for high-resolution image generation, where the training samples are low-density \wrt the high-dimensional feature space. 
Salimans~\etal~\citep{salimans2016improved} presented the popular metric of Inception Score (IS) to measure the individual sample quality. IS does not directly reflect the population-level generation quality, \eg, the overfitting, and loss of diversity. It also requires pre-trained perceptual models on ImageNet or other specific datasets~\citep{barratt2018note}. 
Heusel~\etal~\citep{heusel2017gans} proposed the Fréchet Inception Distance (FID), which models the distribution of image features as multivariate Gaussian distribution and computes the distance between the distribution of real and fakes images. Unlike IS, FID can detect intra-class mode dropping. However, as pointed out by Borji~\etal~\citep{borji2019pros}, the multivariate Gaussian distribution assumption does not hold well on real images, limiting FID's trustworthiness.
Besides IS and FID, Che~\etal~\citep{che2016mode} developed an assessment for both visual quality and variety of samples, known as MODE score, and later shown to be similar to IS~\citep{zhou2017inception}.  
Arora~\etal~\citep{arora2018do, arora2017gans} proposed a test based upon the birthday paradox for estimating the support size of the generated distribution. Although the test can detect severe cases of mode collapse, it falls short in measuring how well a generator captures the true data distribution. It also heavily relies on human annotation, thus hard to scale up to larger-scale evaluation. 
Santurkar~\etal~\citep{santurkar2017classification} and Bau~\etal~\citep{bau2019seeing} are the closet existing works to to our proposed diagnosis. Both approaches took a classification-based perspective and regarded the loss of diversity as a form of covariate shift. Unfortunately, as discussed above, their approaches are ``white box'' and depend on the exposure of original training data. Also, their approaches cannot be extended to subjects without well-defined class labels.

\subsection{Model Calibration for GANs}
There are many efforts addressing the mode collapse problem in GANs.
Some focus on discriminators by introducing different divergence metrics~\citep{metz2016unrolled} and optimization losses~\citep{arjovsky2017wasserstein, mao2017least}.
The minibatch discrimination scheme proposed by Salimans~\etal~\citep{salimans2016improved} allows discrimination between whole mini-batches of samples instead of between individual samples. Durugkar~\etal~\citep{durugkar2016generative} adopted multiple discriminators to alleviate mode collapse. 
Lin~\etal~\citep{lin2018pacgan} proposed PacGAN to mitigate mode collapse by passing $m$ ``packed'' or concatenated samples to the discriminator for joint classification.

ModeGAN~\citep{che2016mode} and VEEGAN~\citep{srivastava2017veegan} enforce the bijection mapping between the input noise vectors and generated images with additional encoder networks. 
Multiple generators~\citep{ghosh2018multi} and weight-sharing generators~\citep{liu2016coupled} are developed to capture more modes of the distribution. However, all of the above assumes (re-)training, and hence are on a different track from our work that focuses on calibrating trained GANs as ``black boxes''.

A handful of works attempted to apply sampling methods to improve GAN generation quality.
Turner~\etal~\citep{turner2018metropolis} introduced the Metropolis-Hastings generative adversarial network (MH-GAN). MH-GAN uses the learned discriminator from GAN training to build a wrapper for the generator for improved sampling at the generation inference stage. With a perfect discriminator, the wrapped generator can sample from the true distribution exactly even with a deficient generator.
Azadi~\etal~\citep{azadi2018discriminator} proposed discriminator rejection sampling (DRS) for GANs, which rejects the generator samples by using the probabilities given by the discriminator to approximately correct errors in the generator's distribution. Nevertheless, these approaches are \textit{white-box} calibration and require access to trained discriminators, which are hardly accessible or even discarded after a GAN is trained.

%% file: chapters/3_Method.tex
\section{Method}\label{sec:method}
\noindent \textbf{Inter-Mode Collapse vs. Intra-Mode Collapse.} Mode collapse happens when there are at least two distant points in the code vector $\mathcal{Z}$ mapped to the same or similar points in the sample space $\mathcal{X}$, whose consequence is limited sample variety in $\mathcal{X}$. There are two distinct concepts here: intra-mode collapse and inter-mode collapse. Inter-mode collapse occurs when some modes (\textit{e.g.}, digit classes in MNIST) are never produced from the generated samples; while intra-mode collapse occurs when all modes (\textit{e.g.}, classes) can be found in the generated samples but with limited variations (\textit{e.g.}, a digit with few writing styles). In this paper, we investigate the \textit{intra-mode collapse} on the task of unconditional GAN image generation, due to its popularity as well as the constraint of missing object labels during generation. Note that all our techniques can be straightforwardly applied to a conditional generation too.

Given an unconditional generator $G$, we can sample an image $\mathcal{I}=G(z)$ by drawing $z$ from a standard Gaussian distribution $\mathbb{N}(z)$. We define that mode collapse happens when the probability of generating samples with a certain condition $f(\mathcal{I}){=}0$ deviates from the expected value of a target distribution. 

For inter-mode collapse, the conditional function $f$ usually specifies the probability of $\mathcal{I}$ belonging to a certain class~\citep{santurkar2017classification,bau2019seeing}.
For \textit{intra-mode collapse}, we favor a conditional function $f$ that can characterize the diversity of samples in a local region. The definition of diversity (loss), especially when it comes to the semantic level, can be elusive and vague. 
To concretize our study subject, we focus on the collapse of the most significant property that makes a generated image ``unique'', \ie, the \textit{identity} (for generated faces, vehicle, \etc). Note that the definition of identity generalizes more broadly than \textit{class}, and can apply to open-set scenarios when the class is not well-defined, such as generating new faces. 
Conceptually, we can measure \textit{intra-mode collapse} \wrt an anchor image $\mathcal{I}^\prime$ by the probability of generating a sample $\mathcal{I}$ with the same identity as $\mathcal{I}^\prime$, \ie, $f(\mathcal{I})=ID(\mathcal{I}^\prime)-ID(\mathcal{I})$. 

\noindent \textbf{Black-Box Setting.} We assume neither access to the original training data, nor the model parameters, nor the class labels of the original data (which might be
inaccessible or even not well defined, as above explained).

\noindent \textbf{Diagnosis \& Calibration.}
Importantly, we never use the identity labels in any form to evaluate sample diversity. Instead, we leverage the embedded features obtained from the deep networks pre-trained for the recognition or re-identification task for subjects such as faces and vehicles. That is based on the known observation that those ``identity'' features can often directly characterize or show strong transferability to depict other essential attributes: \eg, age/gender/race of faces~\citep{savchenko2019efficient} and color/type/brand of vehicles~\citep{zheng2019attributes}. 

Assume we have an identity descriptor $F_{id}(\cdot)$ that produces a unit vector for image $\mathcal{I}$ in the identity embedding space. We can use the identity feature similarity $s(\mathcal{I}^\prime, \mathcal{I})$ between the anchor and sampled images as a probabilistic surrogate to identity matching in our conditional function:
\begin{equation}
f(\mathcal{I}) \sim \mathbb{B}(s(\mathcal{I}^\prime, \mathcal{I})) ,
\end{equation}
where $\mathbb{B}(p)$ denotes Bernoulli distribution with zero-probability $p$.
Thus, the detection metric for our \textit{intra-mode collapse} becomes the expected similarity with the anchor image $\mathcal{I}^\prime$:
\begin{equation}
p(f(\mathcal{I})=0) = \mathbf{E}_{\mathcal{I}}[s(\mathcal{I}^\prime, \mathcal{I})].
\end{equation}
Since only high similarity indicates possible identity matching, we design $s(\cdot, \cdot)$ as a truncated exponential function of inverse feature distance:
\begin{equation}
\label{similarity}
s(\mathcal{I}_1, \mathcal{I}_2) = \frac{1}{e^\theta-1}(e^{max(0,\theta-d(\mathcal{I}_1,\mathcal{I}_2))}-1),
\end{equation}
where $d(\cdot, \cdot)$ is the normalized cosine distance between identity features:
\begin{equation}
\label{distance}
d(\mathcal{I}_1, \mathcal{I}_2)=\frac{1}{\pi} cos^{-1}(<F_{id}(\mathcal{I}_1), F_{id}(\mathcal{I}_2)>) ,
\end{equation}
and $\theta$ is a hyper-parameter specifying the maximum feature distance between two images of the same identity in the embedding space $F_{id}$. Note that $\theta$ only depends on $F_{id}$. We use Monte Carlo sampling to approximate the expected similarity between an anchor image $\mathcal{I}^\prime$ and a randomly sampled image in the generator's learned distribution. We further propose two calibration approaches to alleviate the collapse by ``reshaping the latent space''.

\subsection{Black-box \textit{Intra-Mode Collapse} Diagnosis via Monte Carlo Sampling}
Now we introduce a practical way to evaluate the expected similarity between an anchor image $\mathcal{I}^\prime$ and a randomly sampled image $\mathcal{I}{=}G(z)$:
\begin{equation}
\begin{array}{l}\label{expected_sim}
\mathbf{E}_z[s(\mathcal{I}^\prime, G(z))]=\bigintssss_z s(\mathcal{I}^\prime, G(z))p(z)dz.
\end{array}
\end{equation}
Eq.(\ref{expected_sim}) can be used to quantitatively measure the mode collapse in the neighborhood of the anchor $\mathcal{I}^\prime$ in $G$'s learned distribution. 
We hereby show two extreme cases: Eq.(\ref{expected_sim}) yields $0$ when $G$ produces no images similar to $\mathcal{I}^\prime$ (\ie, $\forall z, d(G(z),\mathcal{I}^\prime) > \theta$); in contrast, 
Eq.(\ref{expected_sim}) yields $1$ when $G$ produces identical images to $\mathcal{I}^\prime$ (\ie, $\forall z, d(G(z),\mathcal{I}^\prime) = 0$).

As the integral in Eq.(\ref{expected_sim}) is generally intractable, we further incorporate \textit{Monte Carlo sampling} to approximate the expectation with the average of $n$ samples from a Gaussian distribution $\{z_i\} \sim \mathbb{N}(z)$:
\begin{equation}
\label{monte}
\mathbf{E}_z[s(\mathcal{I}^\prime, G(z))] \approx \frac{1}{n}\displaystyle \sum_{i=1}^n s(\mathcal{I}^\prime, G(z_i)).
\end{equation}
Considering both scale and normalization in Eq.(\ref{monte}), we further define a metric named \uline{``Monte Carlo-based Collapse Score'' (\textit{MCCS})}:
\begin{equation}
\label{mccs}
\textit{MCCS}(\mathcal{I}^\prime) = 1 / (1-\log(\frac{1}{n}\displaystyle \sum_{i=1}^n s(\mathcal{I}^\prime, G(z_i)))),
\end{equation}
where $n$ is the size of $\mathcal{C}$, a collection of sampled images. \textit{MCCS} ranges between $0$ and $1$: the larger it is, the more $G$ suffers in mode collapse on $\mathcal{I}^\prime$. In Section 4.2, we empirically validate the sampling-efficiency and effectiveness of the proposed \textit{MCCS}.

Finally, the population statistics of \textit{MCCS} can indicate the occurrence of local dense modes in the entire sample space. We use the mean $\mu_\text{mccs}$ and the standard deviation $\sigma_\text{mccs}$ of \textit{MCCS} to quantitatively measure GANs' \textit{intra-mode collapse}:
\begin{equation}
    \begin{split}
        \mu_\text{mccs}=\frac{1}{m}\displaystyle\sum_{i=1}^{m} \textit{MCCS}(G(z_i)), \sigma_\text{mccs}=(\displaystyle\sum_{i=1}^m \frac{(\textit{MCCS}(G(z_i))-\mu_\text{mccs})^2}{m-1})^{1/2}.
    \end{split}
\end{equation}
Note that, to get the population statistics, we need to obtain a collection of sampled anchor images $\mathcal{A}$, whose size is $m$. We have $\mathcal{A} \cap \mathcal{C} = \emptyset$. Details are shown in Section 4.2.1.

\begin{figure}[ht!]
\centering
\begin{tabular}{ccccc}
\subfloat[$|\mathcal{A}|=1k$]{\includegraphics[width = 0.175\textwidth]{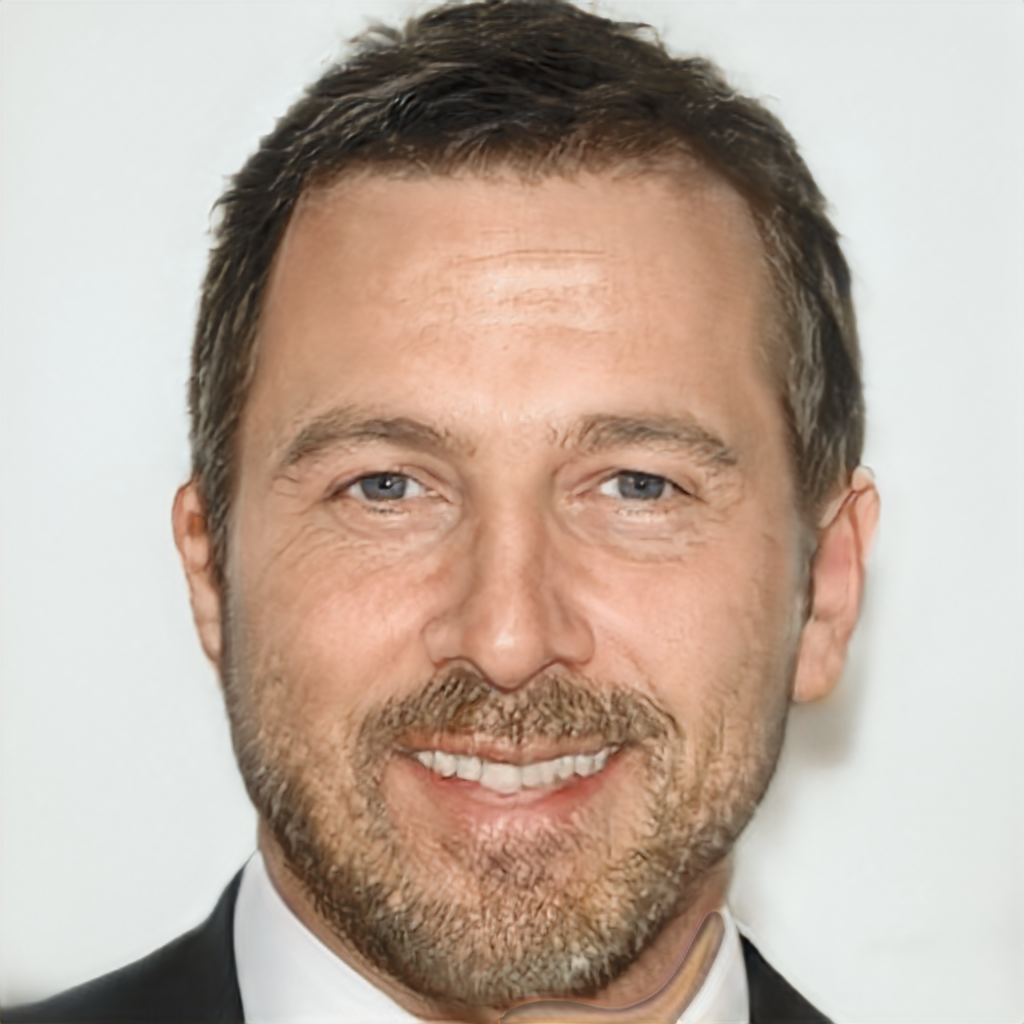}} &
\subfloat[$|\mathcal{A}|=10k$]{\includegraphics[width = 0.175\textwidth]{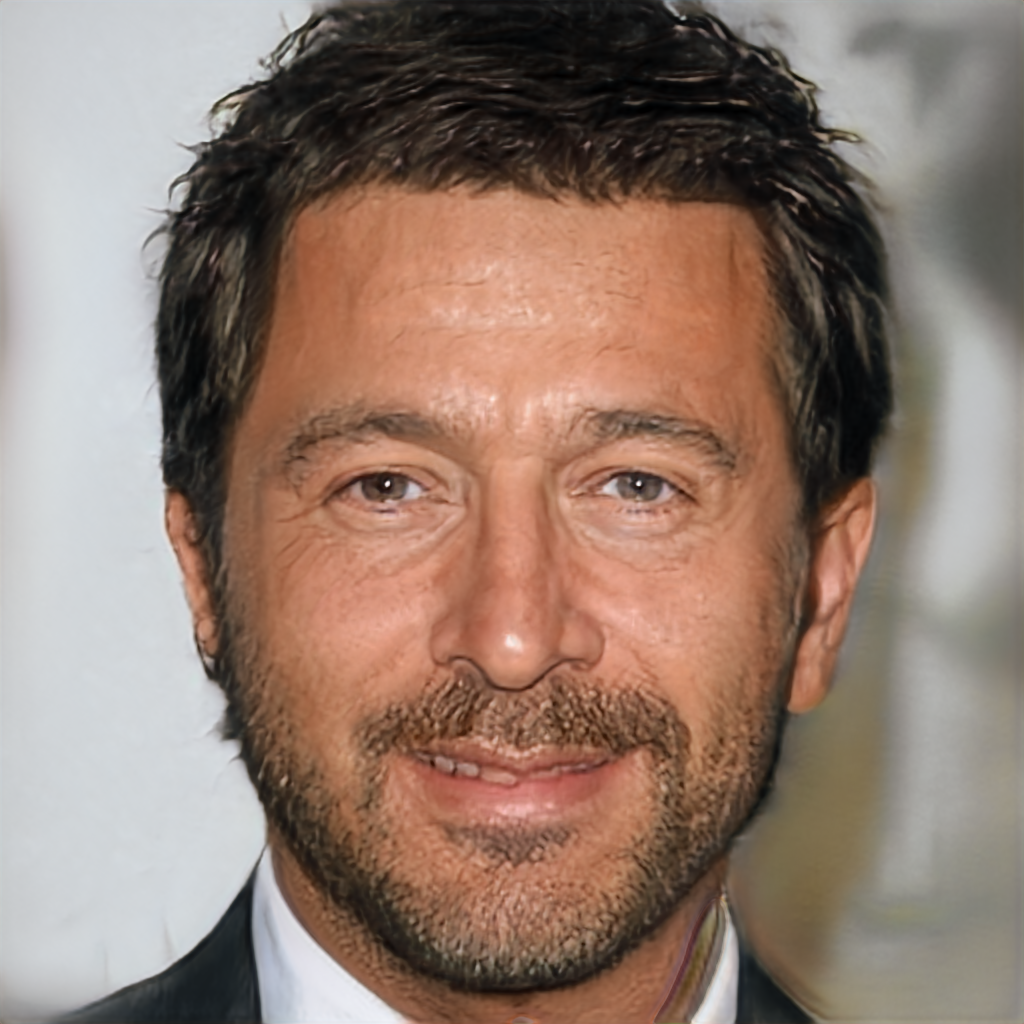}} &
\subfloat[$|\mathcal{A}|=100k$]{\includegraphics[width = 0.175\textwidth]{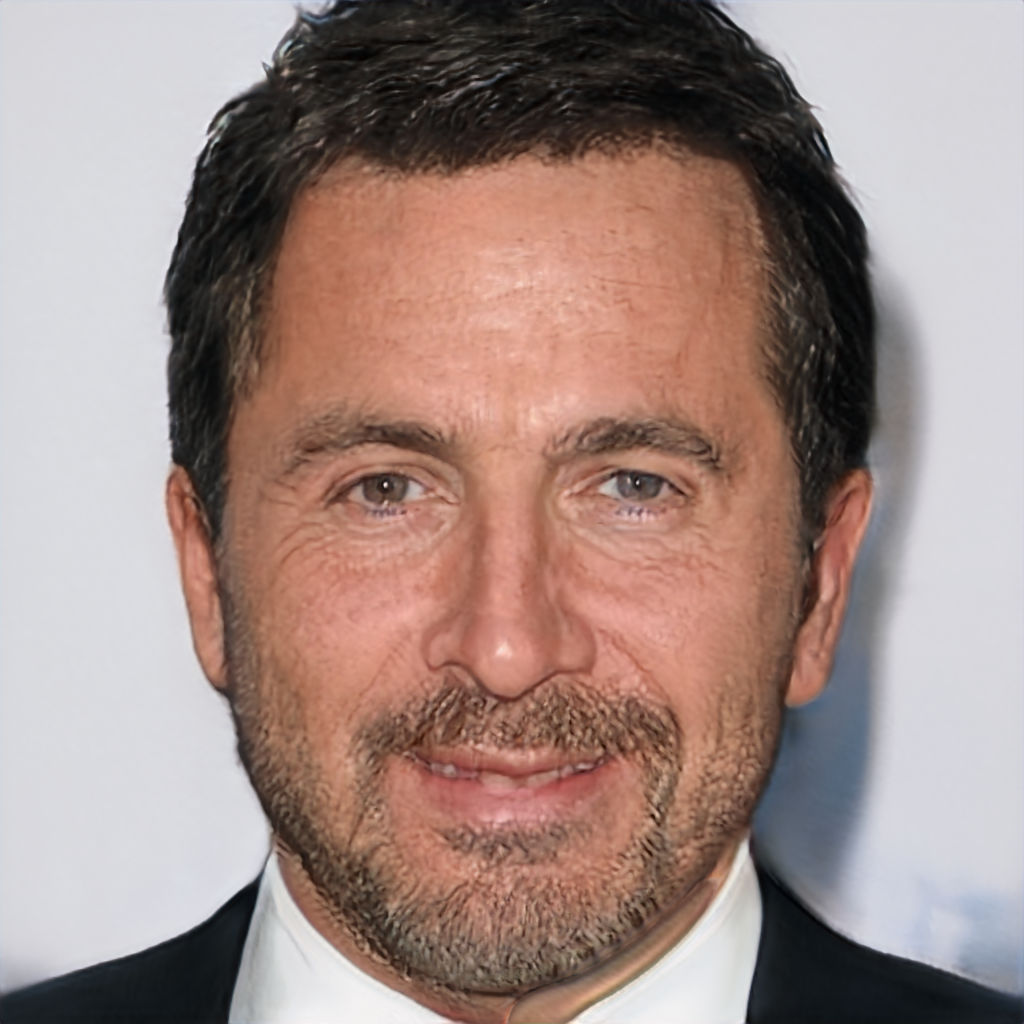}} &
\subfloat[$|\mathcal{A}|=1m$]{\includegraphics[width = 0.175\textwidth]{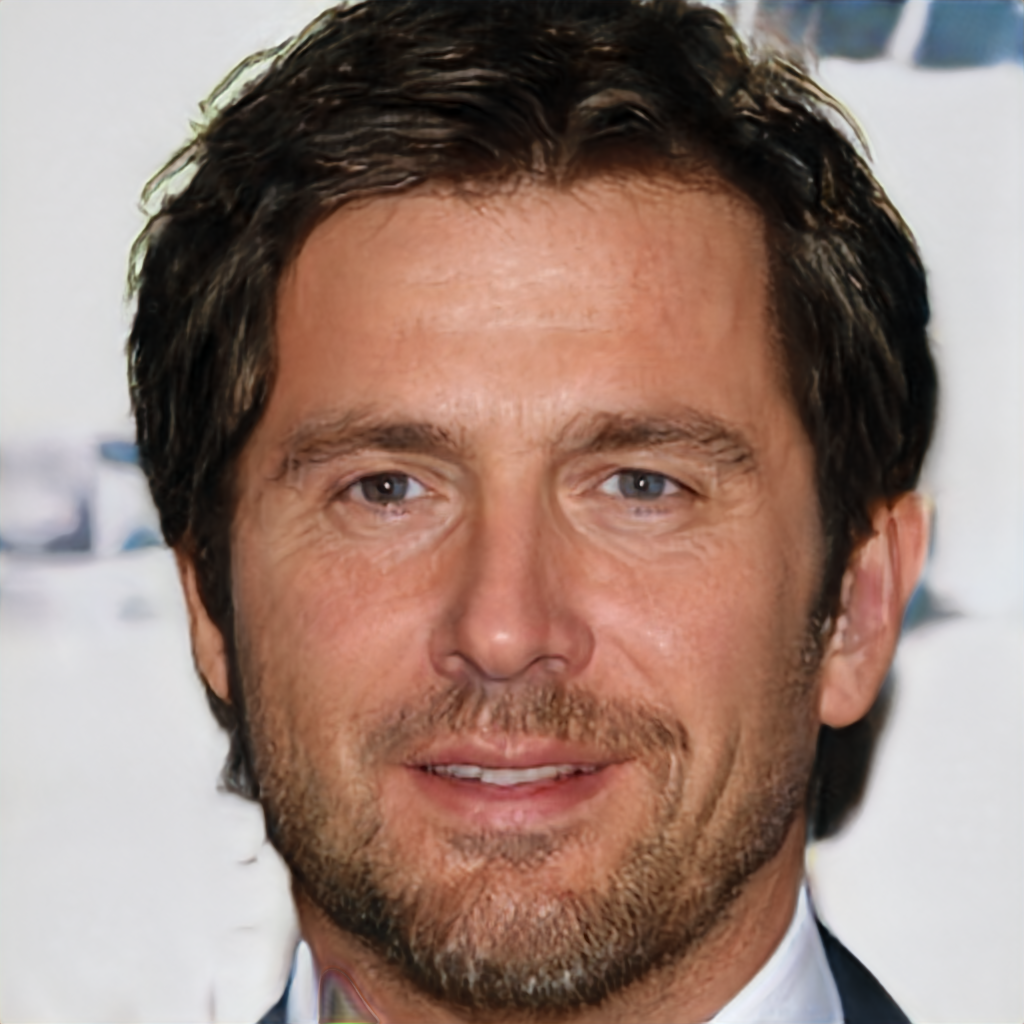}} &
\subfloat[$|\mathcal{A}|=10m$]{\includegraphics[width = 0.175\textwidth]{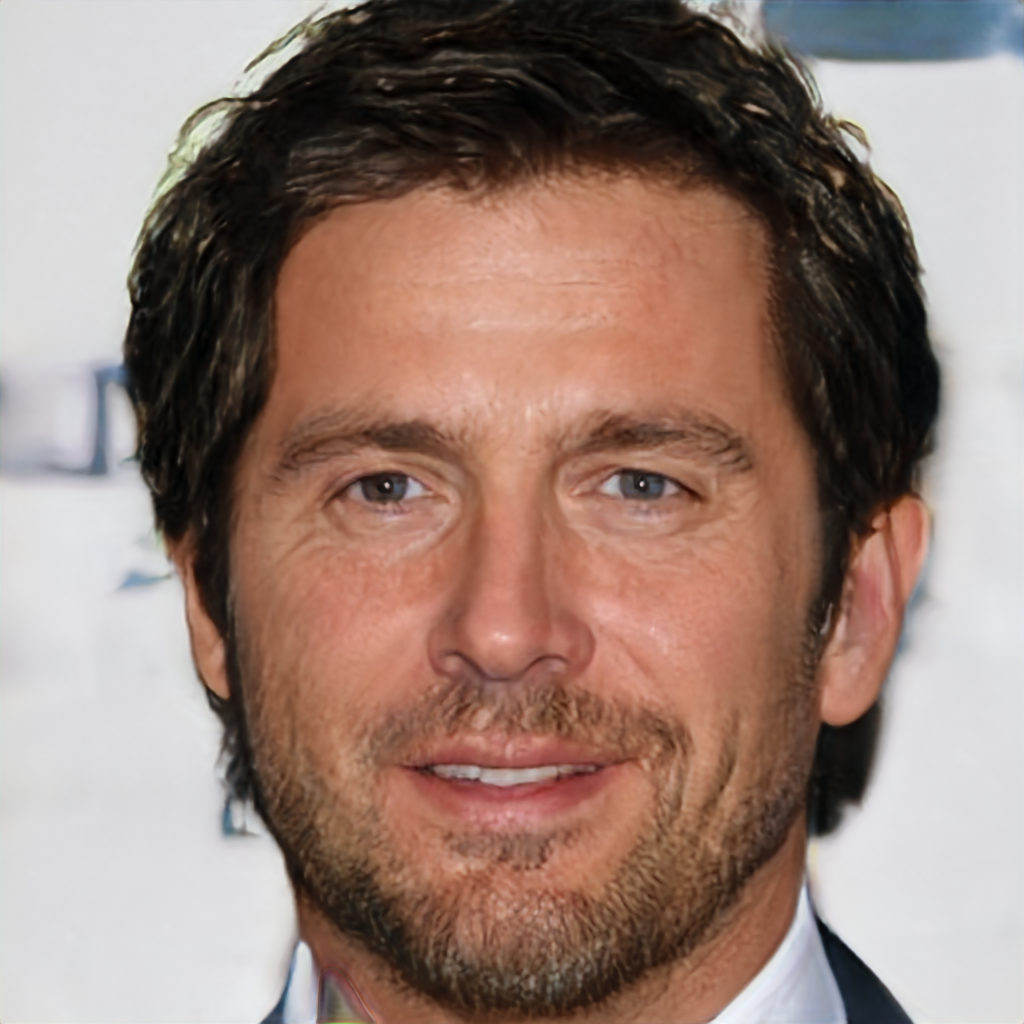}}
\end{tabular}
\caption{Visualization of the \textit{worst-case} dense mode $\mathcal{I}_w$ w.r.t. different size of the $\mathcal{A}$. $\mathcal{A}$ is a collection of randomly sampled anchor images. The $\mathcal{I}_w$ could be reliable obtained when $|\mathcal{A}|=10k$.}
\label{mode_visualization}
\end{figure}

\subsection{Black-box \textit{Intra-Mode Collapse} Calibration via Latent Space Reshaping}
\subsubsection{Calibration \wrt the ``\textit{Worst-Case}" Collapse}
In calibration, we first define the \textit{worst-case} collapsed (dense) mode $\mathcal{I}_w$, \ie, the identity with the largest number of neighbors within a specified distance threshold. Given the radius $r$ of the points neighborhood centered at $\mathcal{I}^\prime$ in the embedding space, a collection of randomly sampled anchor images $\mathcal{A}$, a collection of randomly sampled images $\mathcal{C}$, a worst-case collapsed mode $\mathcal{I}_w$ can be expressed as:
\begin{equation}\label{worst-case}
    \mathcal{I}_w=\argmax_{\mathcal{I}^\prime \in \mathcal{A}} \displaystyle\sum_{\mathcal{I} \in \mathcal{C}} \mathbf{1}_r(d(\mathcal{I}^\prime,\mathcal{I})),
\end{equation}
where $\sum_{\mathcal{I} \in \mathcal{C}} \mathbf{1}_r(d(\mathcal{I}^\prime,\mathcal{I}))$ computes the number of neighbors within $r$ distance of $\mathcal{I}^\prime$ in the embedding space, among all images in $\mathcal{C}$, and $\mathbf{1}_r(\cdot)$ is an indicator function that gives $1$ if $d(\mathcal{I}^\prime,\mathcal{I}) \leq r$.

\begin{wrapfigure}{R}{0.575\textwidth}
\begin{minipage}{0.575\textwidth}
\vspace{-2em}
\begin{algorithm}[H]
\caption{Reshaping Latent Space via Gaussian Mixture Models}
\label{reshape_gmm}
\begin{algorithmic}
\State{$\triangleright$ Given a generator $G$, a neighbor distance threshold $r_0$, and \textcolor{black}{a collection of dense modes $\mathcal{D}$}}
\State{$\triangleright$ $\mathcal{Z}$ $\gets$ $\{z_1, \cdots, z_n\}$} \Comment{$n$ sampled latent codes}
\State{$\triangleright$ $\{\mu_k\}_K, \mathbf{A}_c(\cdot) \gets \text{K-Means}(\mathcal{Z}, K)$} \Comment{Cluster assignment $\mathbf{A}_c: \mathcal{Z} \mapsto \{1,\cdots,K\}$}

\State{$\triangleright$ $w_{sum} \gets 0$} \Comment{The normalization factor}
\ForEach{$k \in \text{range}(1,K)$}
\State{$\triangleright$ $\mathcal{C}_k \gets \{G(z)|\mathbf{A}_c(z)=k,z \in \mathcal{Z}\}$}
\State{\textcolor{black}{$\triangleright$ $w_k \gets 1 / (\sum_{\mathcal{I}_m \in \mathcal{D}}\sum_{\mathcal{I} \in \mathcal{C}_k} \mathbf{1}_{r_0}(d(\mathcal{I}_m,\mathcal{I})))$}} 
\State{$\triangleright$ $w_{sum} \gets w_{sum}+w_k$}
\EndFor
\State{\textbf{return} $\displaystyle\sum_{k=1}^K\frac{w_k}{w_{sum}}\phi(\cdot|\mu_k, \Sigma)$}
\end{algorithmic}
\end{algorithm}
\vspace{-2em}
\end{minipage}
\end{wrapfigure}

We next present two black-box approaches, both focusing on calibrating a detected \textit{worst-case} ``collapsed'' mode $\mathcal{I}_w$.  The calibration aims to maximally alleviate the density of the mode $\mathcal{I}_w$ while preserving the overall diversity and quality of all generated images. 

\noindent \textbf{Biased Anchor Images.} The sampled anchor images could indeed be biased due to limited sampling size. But we have empirically verified that the \textit{worst-case} dense mode $\mathcal{I}_w$ is consistent against sampling. In order to verify the consistency of the \textit{worst-case} dense mode $\mathcal{I}_w$ against sampling, we fix the size of $\mathcal{C}$ to be $1m$ and visualize the $\mathcal{I}_w$ w.r.t. different size of $\mathcal{A}$ in Figure~\ref{mode_visualization}. We consistently observe roughly the same identity as the sampling size increases. Therefore, despite sampling bias in anchor images, $\mathcal{I}_w$ can be reliably obtained even when $|\mathcal{A}|=10k$. The consistency of $\mathcal{I}_w$ demonstrates that the support size of $\mathcal{I}_w$ is nonnegligible. The experiments are conducted on StyleGAN2 trained on CelebAHQ-1024.

\noindent \textbf{Why Calibrating the Worst-Case Mode Only?} We emphasize that our proposed methods can be readily applied to any amounts of collapse modes; however, we have two-fold rationales: (1) focusing on and calibrating the worst-case mode provides good proofs-of-concept and are usually the easiest to demonstrate quantitative and visual gains; (2) we empirically observe that calibrating only the worst-case mode could simultaneously alleviate other collapsed modes, without incurring multiple rounds of sampling overheads. 

\subsubsection{Two Approaches} Given the worst-case dense mode $\mathcal{I}_w$, our proposed calibration approaches alleviate the collapse by ``reshaping the latent space'': they operate on the latent codes as post-processing and require no modification of the trained model nor access to the model parameters or training data, making them completely ``black-box''.

A prerequisite for the proposed calibrations is a \textit{smooth manifold} assumption that comes from empirical observation: as we consistently obtain neighbors that are visually close to $\mathcal{I}_w$, when interpolating near $\mathcal{I}_w$, the latent codes of $\mathcal{I}_w$ are assumed to lay on some smooth manifold. This assumption is mild and well observed in practice.

\noindent \textbf{Approach \#1: Reshaping Latent Space via Gaussian Mixture Models.}
Based on the smooth manifold assumption, the latent space distribution $\phi(z;\xi_0)$ can be approximated with a mixture of Gaussians $\displaystyle\sum_{i=1}^K w_i\phi(z;\xi_i)$. We randomly sample $N$ latent codes and use $K$-means to explore $\xi_k=(\mu_k, \sigma_k)$. We denote $p(\mathcal{I}_w)$ as the probability of sampling the target \textit{worst-case} dense mode $\mathcal{I}_w$:
\begin{equation}
\begin{split}
p(\mathcal{I}_w) &= \bigintssss p(\mathcal{I}_w|z)\phi(z;\xi_0) dz \approx \displaystyle\sum_{k=1}^K w_k \bigintssss p(\mathcal{I}_w|z) \phi_k(z;\xi_k) dz. 
\end{split}
\end{equation}
If $p(\mathcal{I}_w|\xi_k)$ is large, we reduce $w_k$ to make the overall $p(\mathcal{I}_w)$ small. $p(\mathcal{I}_w|\xi_k)$ is estimated by the number of neighbors within $r$ distance to $\mathcal{I}_w$ in the $k_{th}$ cluster $\mathcal{C}_k$, \ie, $\sum_{\mathcal{I} \in \mathcal{C}_k} \mathbf{1}_r(d(\mathcal{I}_w,\mathcal{I}))$. The detailed algorithm is outlined in Algorithm \ref{reshape_gmm}.

\begin{wrapfigure}{R}{0.575\textwidth}
\begin{minipage}{0.575\textwidth}
\vspace{-1.0em}
\begin{algorithm}[H]
\caption{Reshaping Latent Space via Importance Sampling}
\label{reshape_importance}
\begin{algorithmic}
\State{$\triangleright$ Given a generator $G$, a neighbor distance threshold $r_0$, and \textcolor{black}{a collection of dense modes $\mathcal{D}$}}
\State{$\triangleright$ $\mathcal{Z}$ $\gets$ $\{z_1, \cdots, z_n\}$} \Comment{$n$ sampled latent codes}
\State{$\triangleright$ $\mathcal{X} \gets \{G(z_1), \cdots, G(z_n)\}$}
\State{$\triangleright$ $\mathcal{I}_r \gets G(z)$} \Comment{Random image for reference}
\State{\textcolor{black}{$\triangleright$ $\text{IS} \gets \emptyset$}} \Comment{A collection of important sampling parameters}
\ForEach{\textcolor{black}{$\mathcal{I}_m \in \mathcal{D}$}}
\State{$\triangleright$ $p \gets \sum_{\mathcal{I} \in \mathcal{X}} \mathbf{1}_{r_0}(d(\mathcal{I}_r,\mathcal{I})) / \sum_{\mathcal{I} \in \mathcal{X}} \mathbf{1}_{r_0}(d(\mathcal{I}_m,\mathcal{I}))$}
\State{$\triangleright$ $\mathcal{H} \gets \{z|d(G(z), \mathcal{I}_m)\leq r_0, z \in \mathcal{Z}\}$} 
\State{$\triangleright$ $\mathcal{Z}^\prime \gets \{\displaystyle\sum_i \alpha_i z_i:z \in \mathcal{H}, \ \forall i, \alpha_i \geq 0, \ \displaystyle\sum_i\alpha_i=1\}$}
\State{\textcolor{black}{$\triangleright$ $\text{IS} \gets \text{IS} \cup (p,\mathcal{Z}^\prime)$}}
\EndFor
\State{\textbf{return} $\text{IS}$}
\end{algorithmic}
\end{algorithm}
\vspace{-2.0em}
\end{minipage}
\end{wrapfigure}

\noindent \textbf{Approach \#2: Reshaping Latent Space via Importance Sampling.}
Under the same smooth manifold hypothesis, the high-density region corresponding to the target dense mode $\mathcal{I}_w$ can be approximated with a convex hull.

Let the estimated neighboring function densities for the dense and sparse regions be $p_1$ and $p_2$ respectively. We accept the samples from $G$ falling in the high-density region with a probability of $p_2/p_1$ so that the calibrated densities can match. We approximate the high-density region with a convex hull formed by the collection of latent codes corresponding to the identities close to the target dense mode $\mathcal{I}_w$. The details are outlined in Algorithm \ref{reshape_importance}.

Compared to Approach \#1 that relies on Gaussian mixture models, Approach \#2 that relies on importance sampling is often found to be better in preserving the image generation quality. In importance sampling, the high-density region corresponding to the target dense mode $\mathcal{I}_w$ is approximated with a convex hull formed by the collection of the latent codes, whose identity is very close to $\mathcal{I}_w$ in the embedding space. Then, a rejection step is introduced to match the calibrated dense mode with a regular mode. In comparison, in the Gaussian mixture model, there is no explicit formulation of the dense region corresponding to the dense mode $\mathcal{I}_w$.
However, the rejection step based on the explicit formulation of the dense mode via convex hull in the importance sampling approach brings additional computation cost, thus more time-consuming than the mixture model-based approach. We present both options for practitioners to choose from as per their needs.

%% file: chapters/4_Experiments.tex
\section{Experiments}

\begin{wrapfigure}{r}{0.50\textwidth}
    \vspace{-1.0em}
    \includegraphics[width=0.50\textwidth]{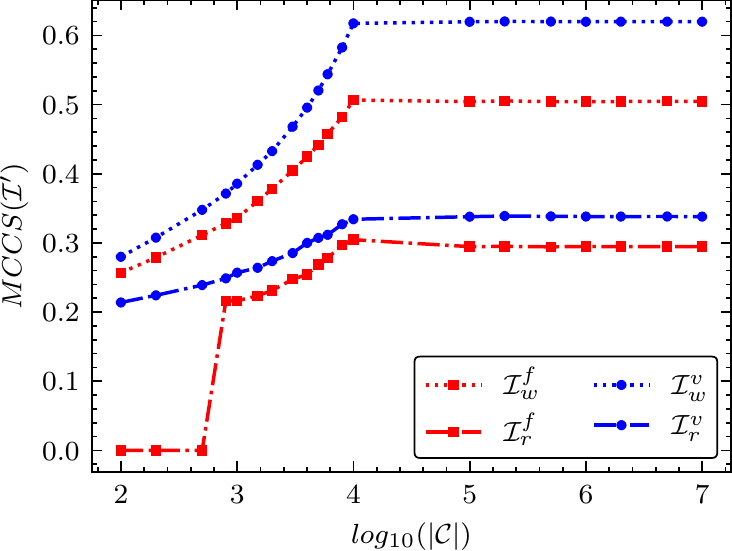}
    \vspace{-2.0em}
    \caption{\small Justifying the sampling-efficiency of $\textit{MCCS}(\mathcal{I}^\prime)$ \wrt $\log_{10}(|\mathcal{C}|)$. $|\mathcal{C}|$ is the number of sampled images. For simplicity, $\mathcal{I}_w$ is the \textit{worst-case} dense mode. $\mathcal{I}_r$ is a \textit{randomly} sampled image. \textit{MCCS} can be reliably obtained at around $|\mathcal{C}|=10^4$. On the face generation task, we have $\textit{MCCS}(\mathcal{I}_w^f)=0.62$ and $\textit{MCCS}(\mathcal{I}_r^f)=0.34$. On the vehicle generation task, we have $\textit{MCCS}(\mathcal{I}_w^v)=0.50$ and $\textit{MCCS}(\mathcal{I}_r^v)=0.30$.}
    \vspace{-2.0em}
    \label{MCCS-sample}
\end{wrapfigure}

\subsection{Experiment Settings}
\subsubsection{Datasets and Models} We choose four state-of-the-art GANs: PGGAN~\citep{karras2017progressive}, StyleGAN~\citep{karras2018style} StyleGAN2~\citep{karras2019analyzing} and BigGAN~\citep{brock2018large}, as our model subjects of study \footnote{While detecting collapse in unconditional GANs is more challenging, our proposed diagnosis can also be directly applied to conditional GANs.}. All are known to be able to produce high-resolution, realistic, and diverse images. The observations below drawn from the four models also generalize to a few other GAN models. We choose high-resolution human face benchmarks of CelebAHQ~\citep{karras2017progressive} and FFHQ~\citep{karras2018style}, and high-resolution vehicle benchmark of LSUN-Car~\citep{yu2015lsun}  as our data subject of study. All benchmarks consist of diverse and realistic images. Lower resolutions are used for fast convergence in training. \Eg, CelebAHQ-128 stands for CelebAHQ downsampled to $128 \times 128$ resolution.

\subsubsection{Choice of $F_{id}$ and Hyperparameters} We use InsightFace~\citep{deng2019retinaface,guo2018stacked,deng2018menpo,deng2018arcface} and RAM~\citep{icme-ram-liu} as $F_{id}$ to serve as the face identity descriptor and vehicle identity descriptor, respectively. We emphasize that the due diligence of ``sanity check'' \footnote{Their face recognition and vehicle re-identification results are manually inspected one-by-one and confirmed to be highly reliable on the generated images. More specifically, dissimilar-looking images are far away from one another in the embedding space, while similar-looking images are close to one another.} has been performed on those classifiers. 


\subsection{Justifying \textit{MCCS}'s Sampling-efficiency and Correctness}

We justify the sampling-efficiency and correctness of \textit{MCCS} in Eq.(\ref{mccs}) on StyleGAN trained on CelebAHQ-$1024$ and BigGAN trained on LSUN-Car-$1024$.

\subsubsection{Sampling-efficiency}
We empirically justify the efficiency of sampling at both sample and population levels.

\noindent \textbf{Sample-level.} We first obtain two pairs of \textit{worst-case} dense mode defined in Eq.(\ref{worst-case}) and a \textit{randomly} sampled image on the face generation task and vehicle generation task respectively, \ie, $(\mathcal{I}_w^f,\mathcal{I}_r^f)$ on face and $(\mathcal{I}_w^v, \mathcal{I}_r^v)$ on vehicle. Then, we compute \textit{MCCS} using a collection of sampled images $\mathcal{C}$ at different sizes. As is shown in Figure \ref{MCCS-sample}, \textit{MCCS} can be reliably obtained at around $|\mathcal{C}|=10^4$ for $(\mathcal{I}_w, \mathcal{I}_r)$ on both face and vehicle generation. 

\begin{wrapfigure}{r}{0.60\textwidth}
    \vspace{-1em}
    \includegraphics[width=0.60\textwidth]{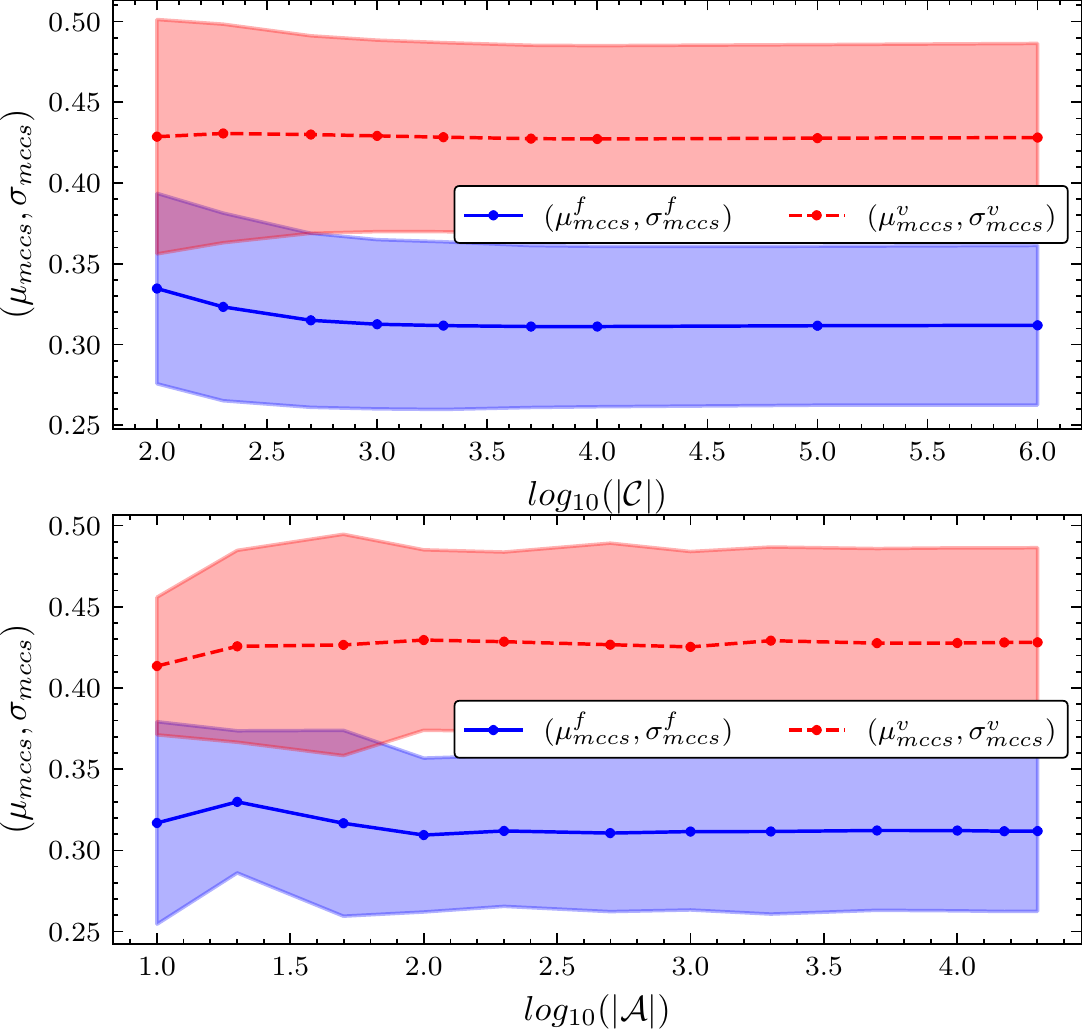}
    \vspace{-2em}
    \caption{\small Justifying the sampling-efficiency of $(\mu_\text{mccs},\sigma_\text{mccs})$ \wrt $\log_{10}(|\mathcal{C}|)$ and $\log_{10}(|\mathcal{A}|)$ on the face generation task and vehicle generation task. $|\mathcal{A}|$ is the number of sampled anchor images. $|\mathcal{C}|$ is the number of sampled images. $\mathcal{A} \cap \mathcal{C} = \emptyset$. $(\mu_\text{mccs},\sigma_\text{mccs})$ can be reliably obtained at around $|\mathcal{A}|=10^4$ and $|\mathcal{C}|=10^4$.}
    \vspace{-1.5em}
    \label{MCCS-population}
\end{wrapfigure}

\noindent \textbf{Population-level.} We first obtain a collection of sampled anchor images $\mathcal{A}$. Then, we draw another collection of sampled images $\mathcal{C}$. Note that $\mathcal{A} \cap \mathcal{C} = \emptyset$. Next, for each anchor image $\mathcal{I}^\prime \in \mathcal{A}$, we compute its value of \textit{MCCS} in $\mathcal{C}$. Finally, we compute $(\mu_\text{mccs}, \sigma_\text{mccs})$ using $\mathcal{A}$ and $\mathcal{C}$ at different sizes. As is shown in Figure \ref{MCCS-population}, $(\mu_\text{mccs}, \sigma_\text{mccs})$ can be reliably obtained at $|\mathcal{C}|=10^4$ and $|\mathcal{A}|=10^4$.

\subsubsection{Correctness}
We empirically justify the correctness of the proposed metric in three experiments: StyleGAN trained on a simulated image set, StyleGAN trained on CelebAHQ, and PacGAN trained on CelebA.

\noindent \textbf{StyleGAN Trained on a Simulated Image Set.} The first experiment is designed to prove that our proposed black-box diagnosis can uniquely detect \textit{intra-mode collapse} cases, when existing evaluation metrics fail to do so. To this end, we curate a new dataset of images, whose ``ground-truth collapses'' are manipulated by us in a fully controlled way. No GAN-generated image is used.

CelebAHQ is a highly imbalanced dataset: among it, $30k$ high-resolution face images of $6,217$ different celebrities, the largest identity class has $28$ images, and the smallest one has only 1. Among the $30k$ faces in CelebAHQ, $20,472$ are White, $4,364$ are Black and $3,154$ are Asian.
Flickr-Faces-HQ Dataset (FFHQ) is another high-quality human face dataset, consisting of $70k$ high-resolution face images, without repeated identities (we manually examined the dataset to ensure so. It is thus ``balanced'' in terms of identity, in the sense that each identity class has one sample. Among the $70k$ faces in FFHQ, $53,481$ are White, $9,381$ are Black, and $7,138$ are Asian.

Since white faces dominate both CelebAHQ and FFHQ, we combine the $30k$ images in CelebAHQ with the $70k$ images in FFHQ, discard the repeated identities, and randomly select $6k$ faces for each race in \{White, Black Asian\}. We called the resulting set \textit{Race-Identity-Calibrated-CelebAFFHQ} \textbf{(RIC)}. Next, we randomly pick one face for each race in the above set, repeat it 1,000 times, and add all repeated faces. This augmented set is called \textit{Race-Identity-Calibrated-CelebAFFHQ-aug} \textbf{(RIC-aug)}. Both RIC and RIC-aug have no inter-mode collapse since the number of different identities are equal across races. However, RIC-aug suffers from strong \textit{intra-mode collapse}.

The two StyleGAN trained on RIC and RIC-aug at resolution of $128$ are denoted as $\mathcal{M}_{RIC}$ (FID=15.36) and $\mathcal{M}_{RIC-aug}$ (FID=15.93). Neither FID or the classification-based study~\citep{santurkar2017classification} could reflect the \textit{intra-mode collapse} in RIC-aug. Our proposed black-box diagnosis can detect \textit{intra-mode collapse} by observing a huge gap between $\textit{MCCS}(\mathcal{I}_w)=0.59$ and $\textit{MCCS}({\mathcal{I}_w^\prime})=0.81$, where $\mathcal{I}_w$ and $\mathcal{I}_w^\prime$ are the \textit{worst-case} dense mode among the generated images of $\mathcal{M}_{RIC}$ and $\mathcal{M}_{RIC-aug}$ respectively.

\begin{wrapfigure}{r}{0.60\textwidth}
	\includegraphics[width=0.60\textwidth]{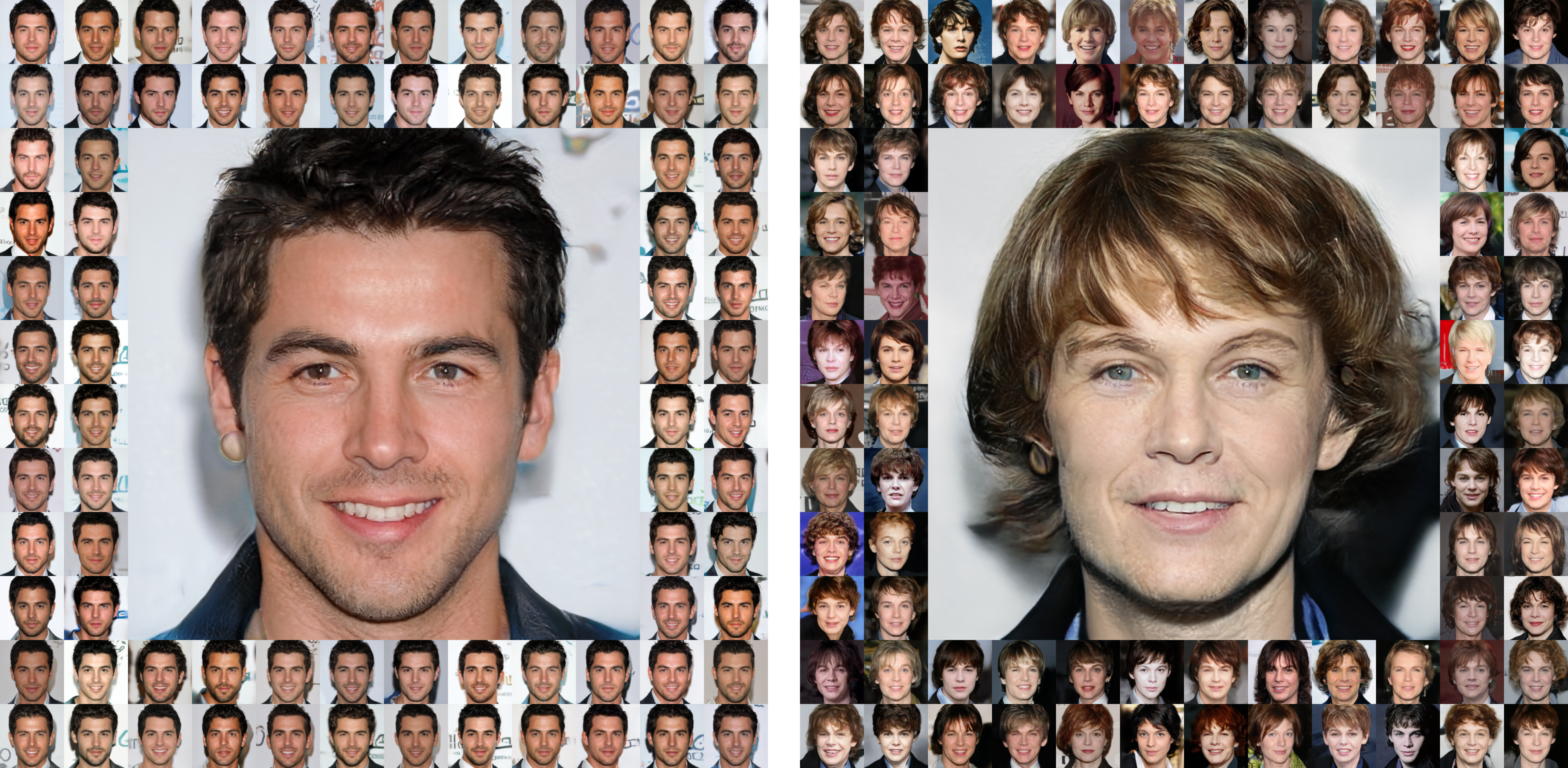}
    \vspace{-2em}
	\caption{\small A ``sanity check'' of \textit{MCCS} is run on StyleGAN trained with CelebAHQ. The left figure is $\mathcal{I}_w$ and the right figure is $\mathcal{I}_r$. Both figures are surrounded with the top $80$ neighbors sorted by distance function in Eq (\ref{distance}).
    }
    \vspace{-2.0em}
	\label{collapse_sanity_check}
\end{wrapfigure}

\noindent \textbf{StyleGAN Trained on CelebAHQ.} Using $|\mathcal{A}|=10^4$ and $|\mathcal{C}|=10^6$ faces sampled from StyleGAN trained on CelebAHQ-$1024$, we run a ``sanity check'' on our proposed \textit{MCCS} and show the results in Figure \ref{collapse_sanity_check}. The left figure $\mathcal{I}_w$, as an anchor face, is the \textit{worst-case} dense mode in Eq.(\ref{worst-case}). The right figure $\mathcal{I}_r$ is a randomly selected anchor face. Both figures are surrounded with the top $80$ neighbors sorted by the distance function defined in Eq (\ref{distance}). $\mathcal{I}_w$ is clearly suffering from mode collapse since all its neighbor are almost identical to it. In contrast, the neighbors of $\mathcal{I}_r$ are quite diverse, though sharing some attribute-level similarities. Importantly, we have $\textit{MCCS}(\mathcal{I}_w, \mathcal{C}) = 0.62$ $\gg$ $\textit{MCCS}(\mathcal{I}_r, \mathcal{C}) = 0.35$, which agrees with the fact that $\mathcal{I}_w$ is the \textit{worst-case} dense mode.

\noindent \textbf{PacGAN Trained on CelebA.}
PacGAN~\citep{lin2018pacgan} also reduces both \textit{inter-} and \textit{intra-mode collapse}s at the same time. It can be represented in the form of ``Pac(X)(m)'', where $X$ is the name of the backbone architecture (\eg, DCGAN~\cite{radford2015unsupervised}), and the integer $m$ refers to the number of samples packed together as input to the discriminator. We conduct our last experiment to run \textit{MCCS} and see if it reflects an improvement. We adopt the SN-DCGAN~\cite{miyato2018spectral} architecture, set the number of packed sample $m$ to be $4$, and train the Pac(SN-DCGAN)(4) on CelebAHQ-128. The FID of SN-DCGAN and Pac(SN-DCGAN)(4) are $34.25$ and $28.12$, respectively. The $(\mu_\text{mccs}, \sigma_\text{mccs})$ of SN-DCGAN and Pac(SN-DCGAN)(4) are $(0.62, 0.18)$ and $(0.54, 0.14)$ respectively. The $\textit{MCCS}(\mathcal{I}_w)$ of SN-DCGAN and Pac(SN-DCGAN)(4) are are $0.84$ and $0.73$ respectively. Our proposed \textit{MCCS} is able to reflect the improvement of PacGAN in both sample and population-level statistics.

\subsection{Black-box Diagnosis on Intra-Mode Collapse}

\subsubsection{Observation of Intra-Mode Collapse on State-Of-The-Art GANs} For StyleGAN (SGAN), StyleGAN2 (SGAN2), PGGAN, and BigGAN (BGAN), despite their observed diversity and high quality in generated images, we still find they are suffering from strong \textit{intra-mode collapse} in Table~\ref{tab:observation_sota}. Note that PGGAN is trained on CelebAHQ-$1024$, StyleGAN on FFHQ-$1024$, StyleGAN2 on LSUN-Car-$1024$, and BigGAN on LSUN-Car-$1024$.

\begin{wraptable}{r}{0.50\textwidth}
\caption{Observation of \textit{intra-mode collapse} on state-of-the-art GANs.} 
\resizebox{0.50\textwidth}{!}{%
\centering
\begin{tabular}{c|c|c|c|c} 
\hline
{} & SGAN & PGGAN &  SGAN2 & BGAN \\
\hline
$(\mu_\text{mccs}, \sigma_\text{mccs})$ & (0.41, 0.06) & (0.48, 0.07)  & (0.31, 0.05)  & (0.45, 0.07)\\
\hline
$\textit{MCCS}(\mathcal{I}_w)$ & 0.64  & 0.72   & 0.56  & 0.69 \\
\hline
\end{tabular}
\label{tab:observation_sota}
}
\end{wraptable}

\subsubsection{Empirical Study on the Cause of Intra-Mode Collapse}
We hypothesize multiple factors that may potentially lead to the observed dense mode (indicating \textit{intra-mode collapse}) of face identity. We perform additional experiments, aiming to validate one by one. Despite the variance for the obtained sample-level and population-level statistics on \textit{MCCS}, \ul{none of them} was observed to cause the observed mode collapse. That implies the existence of some more intrinsic reason for the mode collapse in GAN, which we leave for future exploration.

\noindent \textbf{Imbalance of Training Data?}
CelebAHQ is a highly imbalanced dataset: among its $30k$ high-resolution face images of $6,217$ different celebrities. It is natural to ask: would a balanced dataset alleviate the mode collapse? We turn to the Flickr-Faces-HQ Dataset (FFHQ), a high-quality human face dataset created in~\citep{karras2018style}, consisting of $70k$ high-resolution images, without repeated identities. FFHQ dataset does not have an imbalance in facial attributes.
To further eliminate the attribute-level imbalance, \eg, race, and gender, we combine the $30k$ images in CelebAHQ with the $70k$ images in FFHQ, and discard repeated images in identities. When selecting $6k$ faces for each race in \{White, Black Asian\}, we intentionally make the resulting set balanced in gender for each race. The resulting set is dubbed as Gender-Race-Identity-Calibrated-CelebAFFHQ (GRIC).
As shown in Table~\ref{tab:data_imba}, while the generation quality of StyleGAN trained on GRIC is still high, the \textit{intra-mode collapse} persists and seems to be no less than StyleGAN on CelebAHQ and FFHQ. Therefore, imbalance of training data, regardless of at attribute-level or identity-level, does not cause \textit{intra-mode collapse}.

\begin{table}[!htb]
\centering
\tiny
\begin{minipage}[t]{.44\linewidth}
\caption{Empirical study on the cause of \textit{intra-mode collapse}: imbalance of training data?} 
\centering
\begin{tabular}{c|c|c|c} 
\hline
{} & CelebAHQ & FFHQ & GRIC \\
\hline
$(\mu_\text{mccs}, \sigma_\text{mccs})$ & (0.44, 0.08) & (0.41, 0.06) & (0.43, 0.06) \\
\hline
$\textit{MCCS}(\mathcal{I}_w)$ & 0.67 & 0.64 & 0.62 \\
\hline
\end{tabular}
\label{tab:data_imba}
\end{minipage}%
\hfill\hfill
\begin{minipage}[t]{.55\linewidth}
\caption{Empirical study on the cause of \textit{intra-mode collapse}: model architecture differences?} 
\begin{tabular}{c|c|c|c|c|c} 
\hline
{} & {} & 128 & 256 & 512 & 1024 \\
\hline
\multirow{2}{*}{SGAN} & $(\mu_\text{mccs}, \sigma_\text{mccs})$ & (0.43,0.06) & (0.44,0.06) & (0.43,0.06) & (0.42,0.06) \\
\cline{2-6}
& $\textit{MCCS}(\mathcal{I}_w)$ & 0.64 & 0.63 & 0.65 & 0.65 \\
\hline
\multirow{2}{*}{PGGAN} & $(\mu_\text{mccs}, \sigma_\text{mccs})$ & (0.52,0.08) & (0.51,0.09) & (0.53,0.08) & (0.52,0.08) \\
\cline{2-6}
& $\textit{MCCS}(\mathcal{I}_w)$ & 0.74 & 0.77 & 0.78 & 0.75 \\
\hline
\end{tabular}
\label{tab:arch_diff}
\end{minipage}
\vspace{-1.5em}
\end{table}

\noindent \textbf{Randomness during Optimization?}
We repeat training StyleGAN on CelebAHQ-128 for 10 times, with different random initializations and mini-batching. We have $(\mu_\text{mccs}, \sigma_\text{mccs})$ consistently around $(0.43, 0.06)$. Despite little variances, the \textit{intra-mode collapse} persists. We conclude that randomness during optimization is not the reason for \textit{intra-mode collapse}.

\noindent \textbf{Unfitting/Overfitting in Training?}
We train StyleGAN on CelebAHQ-128 again, and store model checkpoints at iteration $7707$ (FID = $7.67$, same hereinafter), $8307$ ($7.02$), $8908$ ($6.89$), $9508$ ($6.63$), $10108$ ($6.41$), and $12000$ ($6.32$). As the training iterations increases, $(\mu_\text{mccs}, \sigma_\text{mccs})$ decreases from $(0.51, 0.09)$ to $(0.43, 0.06)$. Therefore, the \textit{intra-mode collapse} persists, regardless of stopping the training earlier or later. We point out that unfittinng or overfitting is not the cause of \textit{intra-mode collapse}.

\begin{wrapfigure}{l}{0.50\textwidth}
\vspace{-1em}
	\includegraphics[width=0.50\textwidth]{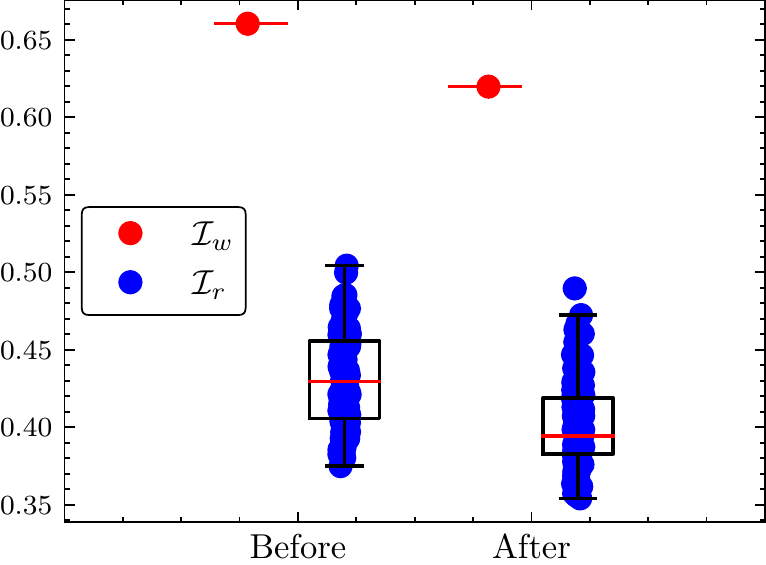}
\caption{\small Reshaping Latent Space via Gaussian Mixture Models: mode collapse analysis of StyleGAN on CelebAHQ-128, before/after Gaussian mixture model-based reshaping.}
\vspace{-3.5em}
\label{gmm_reshaping}
\end{wrapfigure}

\noindent \textbf{Model Architecture Differences?}
Both StyleGAN and PGGAN progressively grow their architectures to generate images of different resolutions: $128$, $256$, $512$, and $1024$. Thus, we train StyleGAN and PGGAN on CelebAHQ-128, CelebAHQ-256, CelebAHQ-512, and CelebAHQ-1024, respectively. According to Table~\ref{tab:arch_diff}, varying the architectures does not eliminate the \textit{intra-mode collapse} either. Thus, we empirically show that different model architectures do not lead to \textit{intra-mode collapse}.

\subsubsection{Diagnosis on Other Fine-Grained Image Generation} Flowers-$102$ consists of $102$ flower categories and is divided into $2,040$ images for training and $6,149$ for testing. CUB-$200$ has $200$ bird categories and is split into $5,994$ images for training and $5,794$ images for testing. We denote the identity descriptor for flower and bird as $F_{id}^f$ and $F_{id}^b$, respectively. Similarly, we denote the image generator for flower and bird as $G^f$ and $G^b$.

To obtain an accurate flower identity descriptor $F_{id}^f$, we adopt the EfficientNet~\cite{tan2019efficientnet} and pretrain it on LifeCLEF2021 Plant Identification~\cite{goeau2020overview,plantclef2021,goeau2016plant,goeau2017plant}. Later on, we finetune it on a curated dataset that combines Jena Flowers 30~\cite{DVN/QDHYST_2017}, Flowers Recognition~\cite{flowers2016kaggle}, Flowers~\cite{flowers2020roboflow}, and Flowers-$17$ \& Flowers-$102$~\cite{flowers17and102}. As an image generator, MSG-GAN~\cite{karnewar2020msg} is trained on Flowers-$102$'s union of training and testing set, a total of $8,189$ images.

\begin{wraptable}{r}{0.75\textwidth}
\caption{Experiments on CUB-200 and Flowers-102} 
\resizebox{0.75\textwidth}{!}{
\begin{tabular}{c c c c c} 
\hline 
Dataset & Image Generator ($G$) & Identity Descriptor ($F_{id}$) & $(\mu_\text{mccs}, \sigma_\text{mccs})$ & $\textit{MCCS}(\mathcal{I}_w)$\\ [0.5ex] 
\hline 
Flowers-$102$~\cite{nilsback2008automated} & MSG-GAN~\cite{karnewar2020msg} & EfficientNet~\cite{tan2019efficientnet} & $(0.69,0.12)$ & $0.84$ \\ 
CUB-$200$~\cite{WelinderEtal2010} & StackGAN-v2~\cite{zhang2018stackgan++} & API-NET~\cite{zhuang2020learning} & $(0.64,0.09)$ & $0.86$ \\ [1ex] 
\hline 
\end{tabular}
\label{table:bird_flower} 
}
\end{wraptable}

To obtain an accurate bird identity descriptor $F_{id}^b$, we adopt the API-NET~\cite{zhuang2020learning} and pretrain it on the DongNiao International Birds $10000$ (DIB-$10$K)~\cite{mei2020dongniao}. Later on, we finetune it on a curated dataset that combines Bird265~\cite{bird265kaggle} and NABirds~\cite{van2015building}. As an image generator, StackGAN-v2 is trained on CUB-$200$, a total of $11,788$ images.

According to Table~\ref{table:bird_flower}, on the fine-grained image generation of birds and flowers, despite observed diversity and high quality in generated images, we can still spot strong intra-mode collapse.

\subsection{Black-box Calibration on Intra-Mode Collapse}
\subsubsection{Reshaping Latent Space via Gaussian Mixture Models}
Starting from a StyleGAN model $\mathcal{M}$ pre-trained on CelebAHQ-128, we aim at alleviating the collapse on the \textit{worst-case} dense mode $\mathcal{I}_w$. We reshape the latent space of $\mathcal{M}$ via Gaussian mixture models to get the new model $\mathcal{M}'$. We get the new \textit{worst-case} dense mode $\mathcal{I}_w'$ in $\mathcal{M}'$. The $(\mu_\text{mccs}, \sigma_\text{mccs})$ has decreased from $(0.43, 0.06)$ to $(0.41, 0.05)$. The $\textit{MCCS}(\mathcal{I}_w)$ has also decreased from $0.66$ to $0.61$. Such an alleviation is achieved with an unnoticeable degradation of generation quality, with FID increasing from 5.93 ($\mathcal{M}$) to 5.95 ($\mathcal{M}'$). As is shown in Figure \ref{gmm_reshaping}, after applying the latent space reshaping, the \textit{intra-mode} collapse has been alleviated, which are indicated by the large gap between $\mathcal{I}_w$ (before) and $\mathcal{I}_w^\prime$ (after), and the large gap between $\mathcal{I}_r$ (before) and $\mathcal{I}_r^\prime$ (after). For the sake of readability and visual quality, in each boxplot, only $100$ randomly chosen $\mathcal{I}_r$ are shown.

\subsubsection{Reshaping Latent Space via Importance Sampling}

\begin{wrapfigure}{r}{0.50\textwidth}
\vspace{-1em}
	\includegraphics[width=0.50\textwidth]{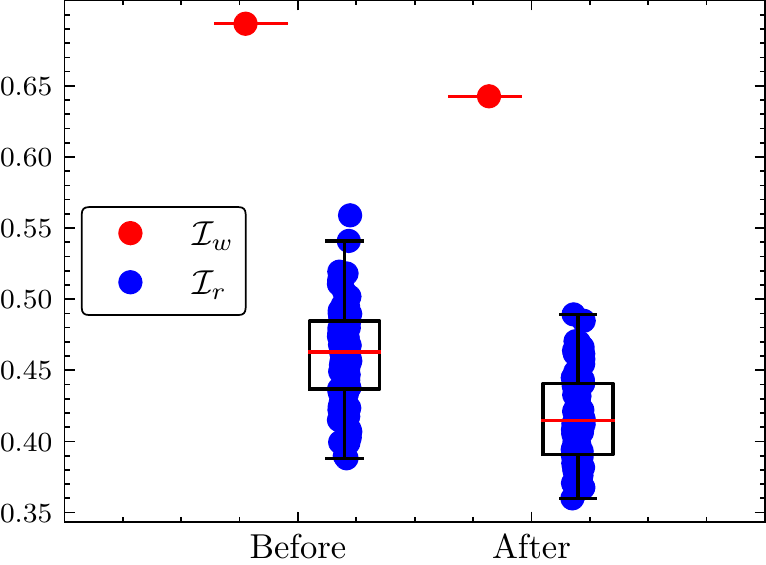}
	\vspace{-2em}
\caption{\small Reshaping Latent Space via Importance Sampling: mode collapse analysis of PGGAN on FFHQ-128, before/after importance sampling-based reshaping.}
\label{is_reshaping}
\vspace{-1.5em}
\end{wrapfigure}

The experiment setting is mostly similar to the reshaping latent space via the Gaussian mixture models case, except that we are using PGGAN trained on FFHQ-128. We integrate importance sampling to the latent code generation stage. Given the dense mode $\mathcal{I}_w$, we can find the collection of latent codes from the top $10^2$ latent codes whose corresponding images have the smallest distances Eq.(\ref{distance}) to $\mathcal{I}_w$, among the $10^6$ random samples. We get the new \textit{worst-case} dense mode $\mathcal{I}_w'$ in $\mathcal{M}'$. The $(\mu_\text{mccs}, \sigma_\text{mccs})$ has decreased from $(0.46, 0.07)$ to $(0.42, 0.06)$. The $\textit{MCCS}(\mathcal{I}_w)$ has also decreased from $0.69$ to $0.64$. The \textit{intra-mode collapse} is again alleviated without sacrificing the visual quality of generated images, since FID only marginally increases from 9.43 ($\mathcal{M}$) to 9.46 ($\mathcal{M}'$). As is shown in Figure \ref{is_reshaping}, after applying the latent space reshaping, the \textit{intra-mode} collapse has been alleviated, which are indicated by the large gap between $\mathcal{I}_w$ (before) and $\mathcal{I}_w^\prime$ (after), and the large gap between $\mathcal{I}_r$ (before) and $\mathcal{I}_r^\prime$ (after). For the sake of readability and visual quality, in each boxplot, only $100$ randomly chosen $\mathcal{I}_r$ are shown.

\subsubsection{Why Calibrating the $\mathcal{I}_w$ Could Benefit the Calibration of Other Modes?}
As is shown in Figure~\ref{fig:visualization}, we visualize the top $24$ modes with the largest number of neighbors within $0.25$ distance and found that they look very similar. Thus, we conclude that the dense region corresponding to $\mathcal{I}_w$ has occupied a considerably large portion in the number of supports in the GAN's learned face identity distribution. Calibrating the \textit{worst-case mode} $\mathcal{I}_w$ implicitly takes the entire dense region into account, since the entire dense region corresponding to the same face identity of $\mathcal{I}_w$.

%% file: chapters/5_Limitations.tex
\section{Discussions and Future Work}
This paper is intended as a pilot study on the intra-mode collapse issue of GANs, under a novel and hardly explored black-box setting. Using face and vehicle as study subjects, we quantify the general intra-mode collapse via statistical tools, discuss and verify possible causes, as well as propose two black-box calibration approaches for the first time to alleviate the mode collapse. Despite the preliminary success, the current study remains to be limited in many ways. First, there are inevitably prediction errors for the identity description on generated images, and even we have done our best to use the most accurate descriptors. Moreover, the fundamental causes of GAN mode collapse call for deeper understandings. We hope our work to draw more attention to studying both the intra-mode collapse problem and the new black-box setting.

\begin{figure}[htb!]
\setlength{\tabcolsep}{0em}
\centering
\begin{tabular}{cccccccc}
		\includegraphics[width=0.075\textwidth]{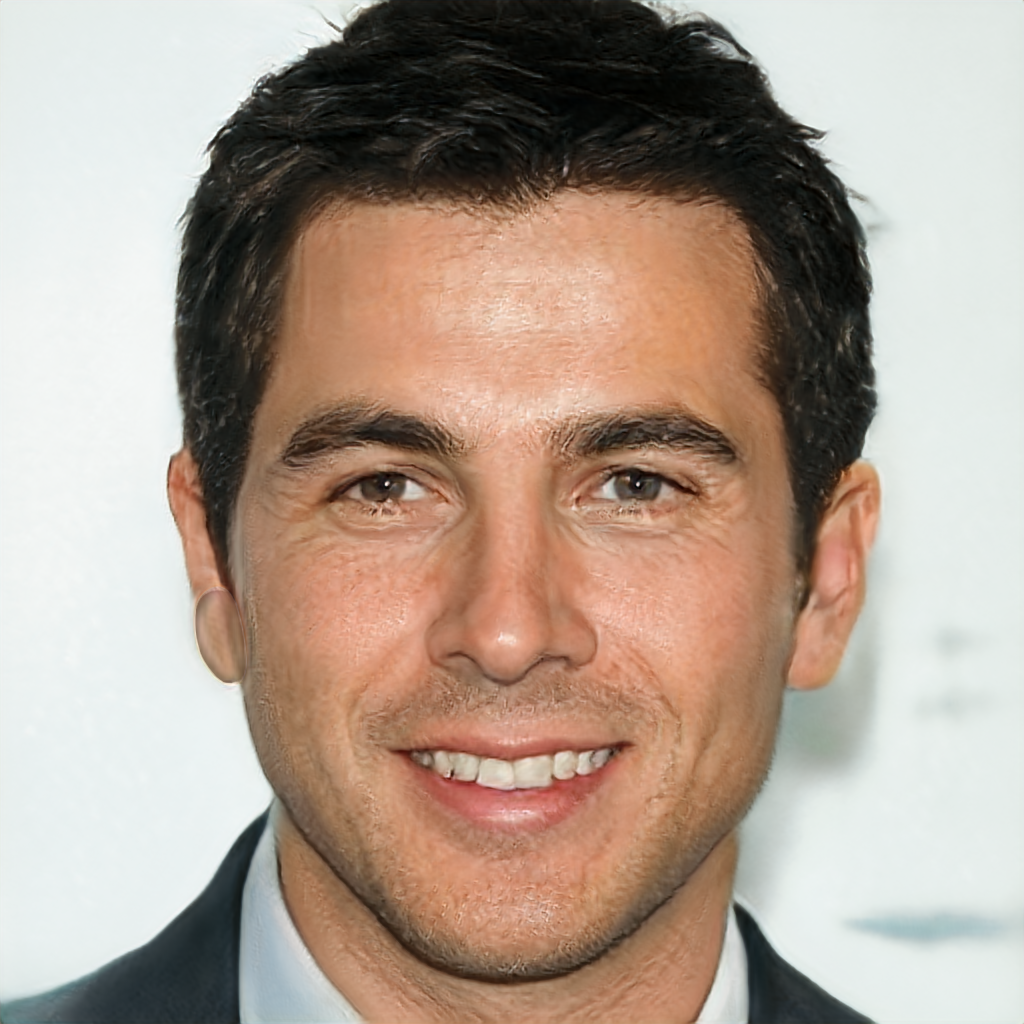}
		\includegraphics[width=0.075\textwidth]{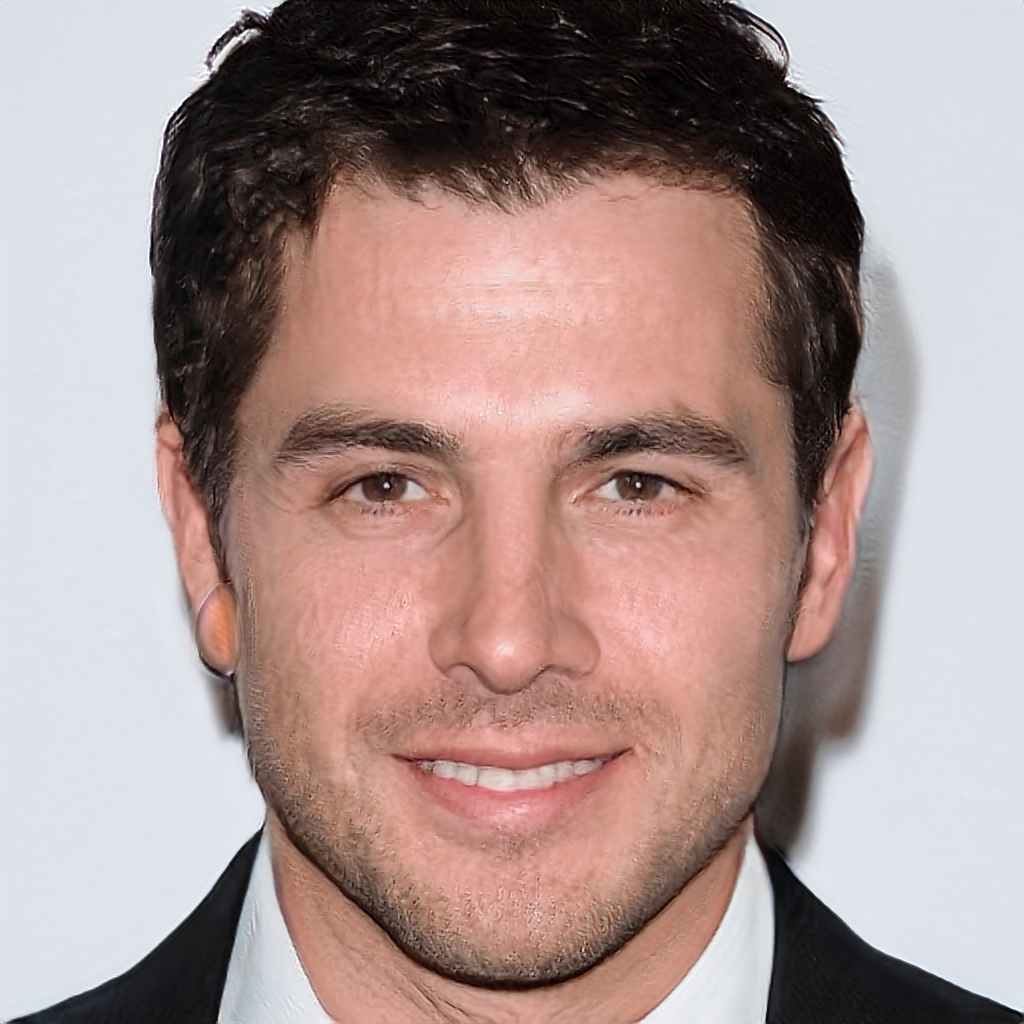}
		\includegraphics[width=0.075\textwidth]{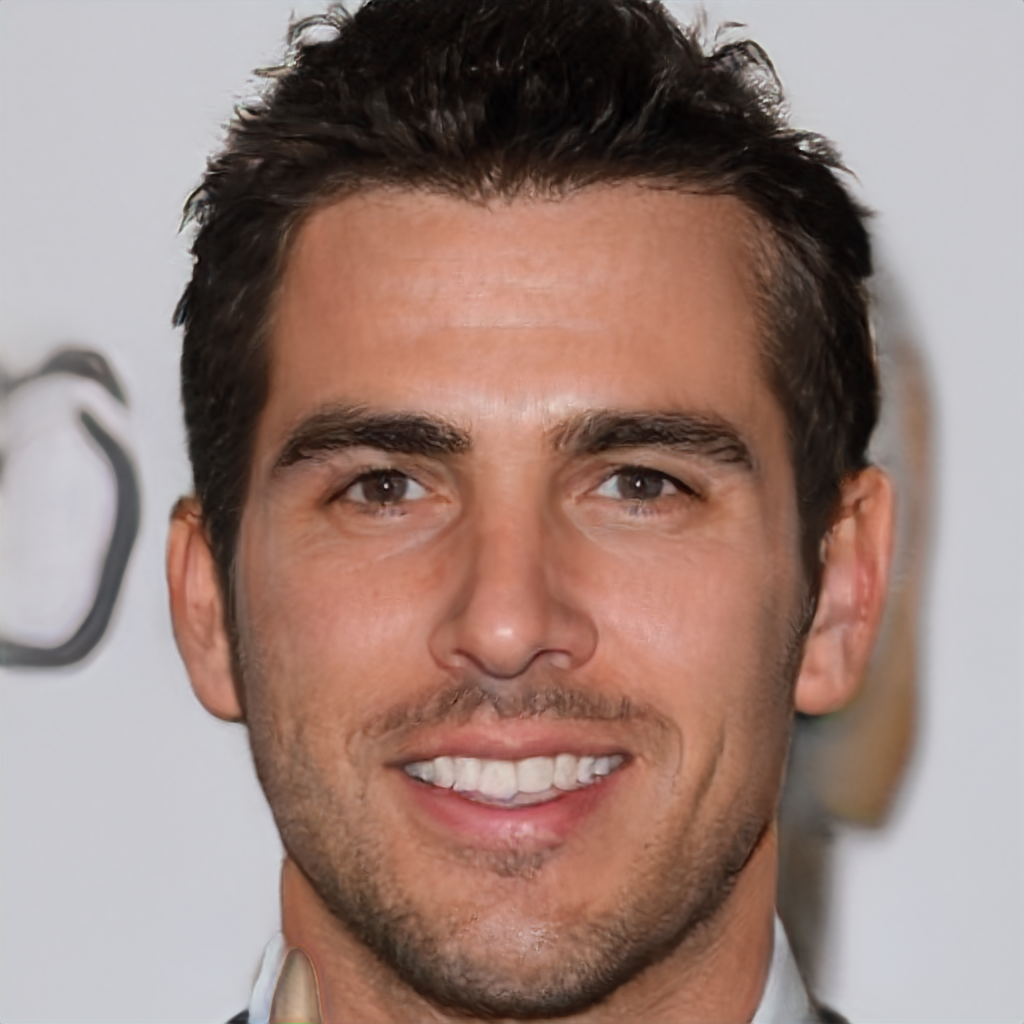}
		\includegraphics[width=0.075\textwidth]{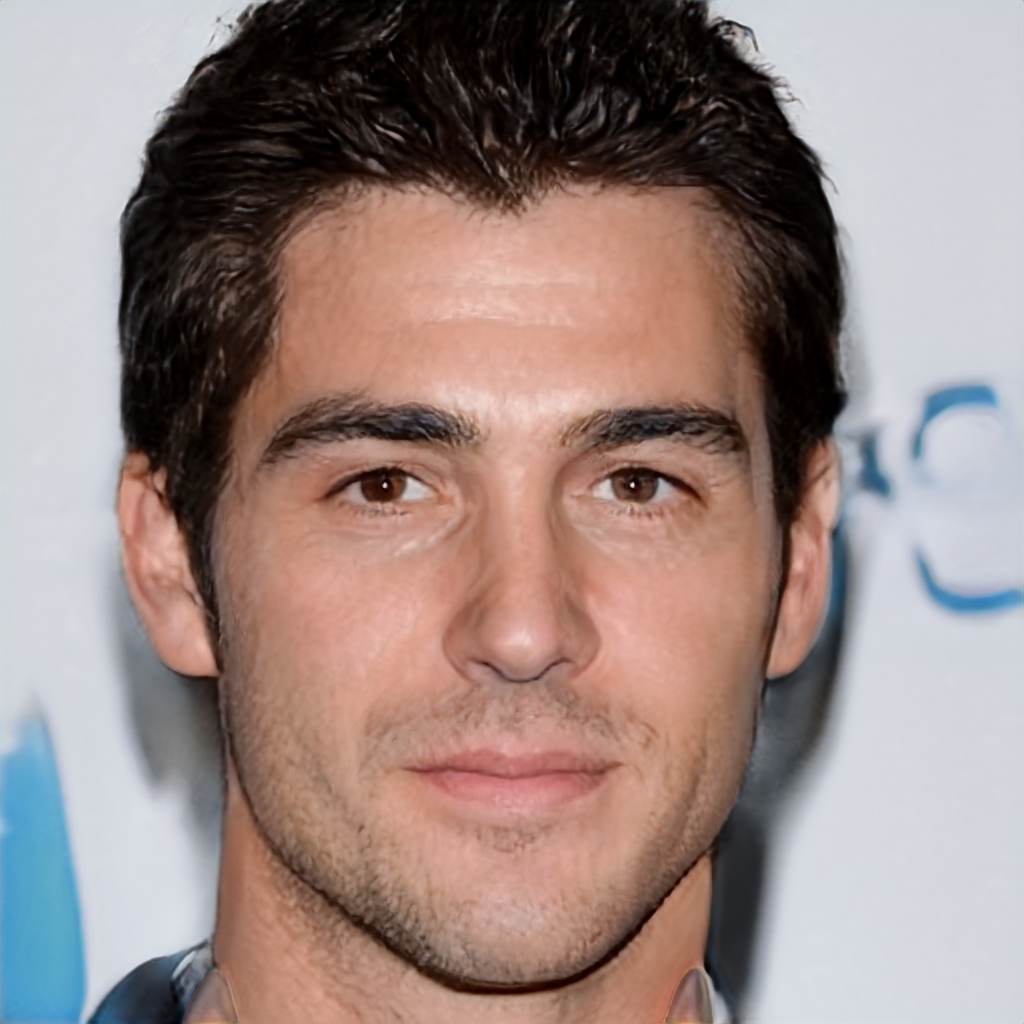}
		\includegraphics[width=0.075\textwidth]{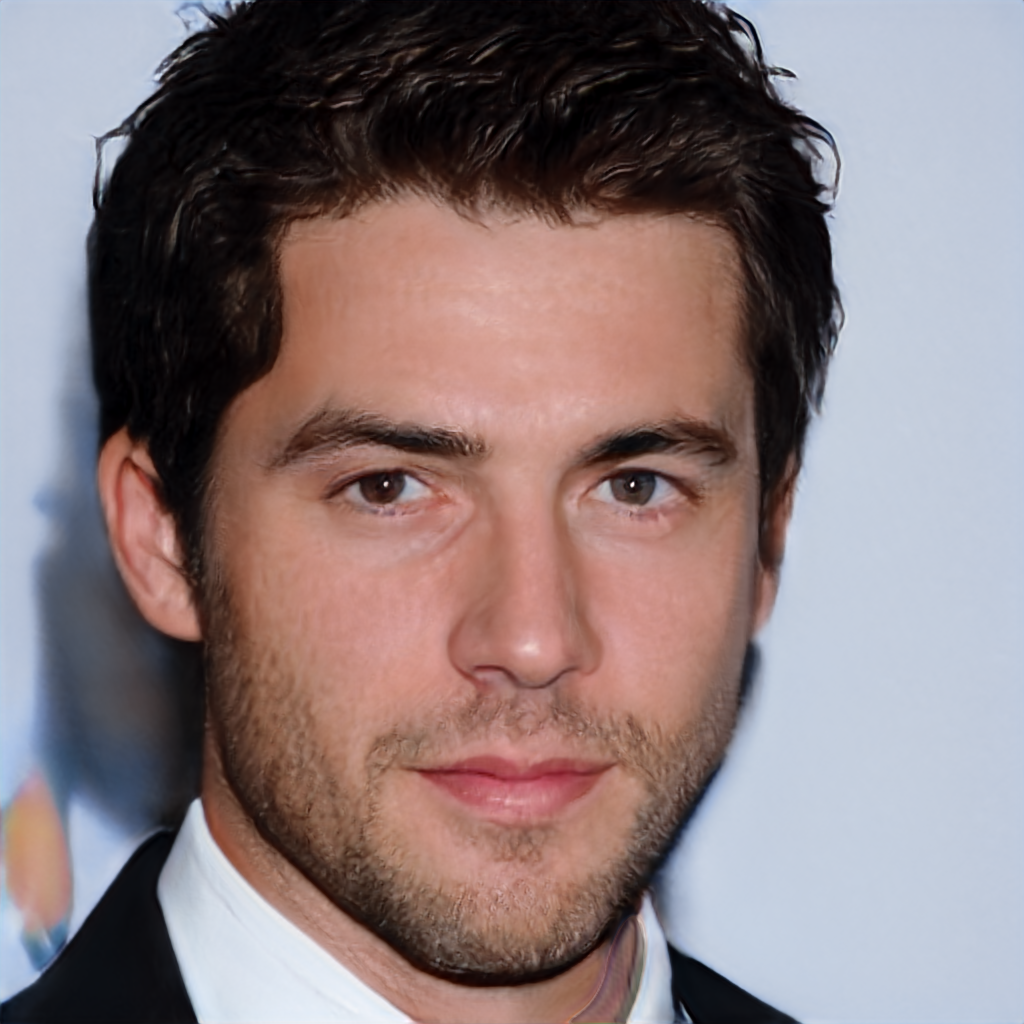}
		\includegraphics[width=0.075\textwidth]{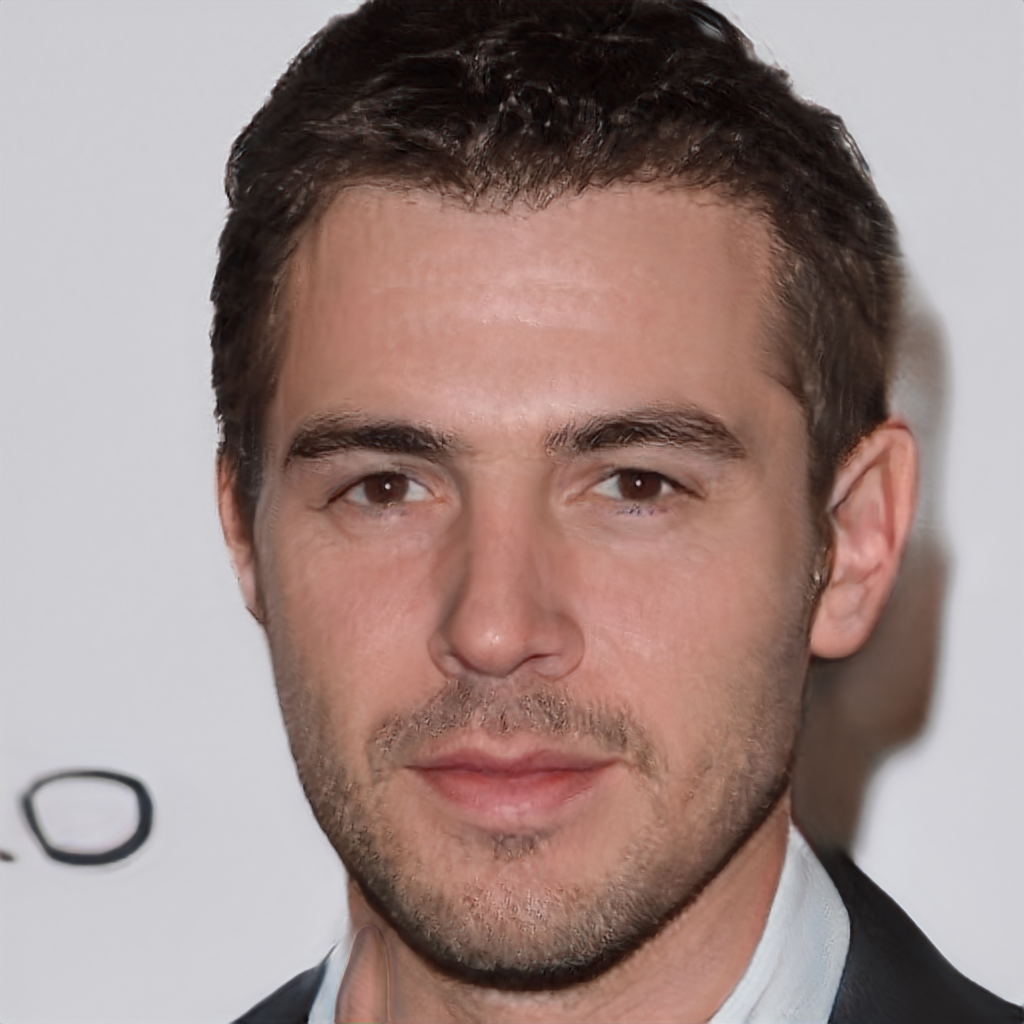}
		\includegraphics[width=0.075\textwidth]{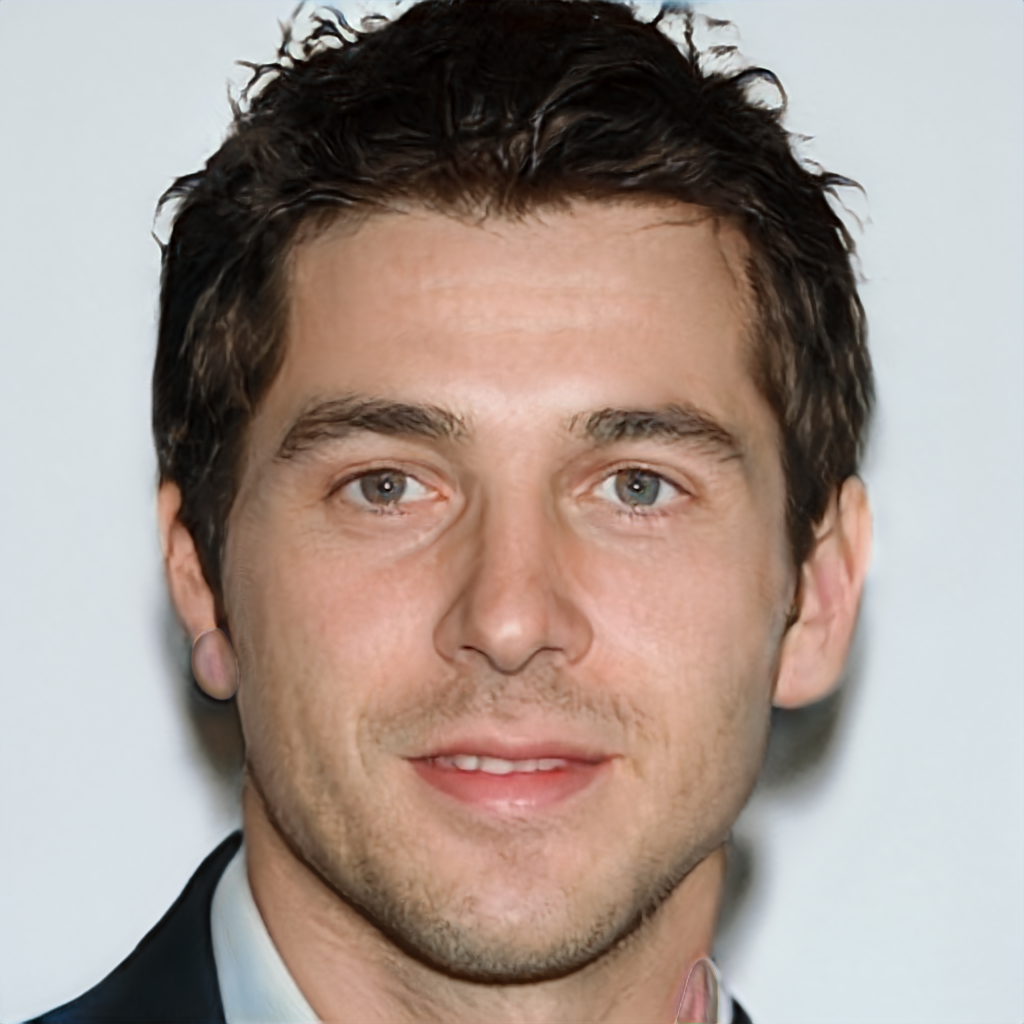}
		\includegraphics[width=0.075\textwidth]{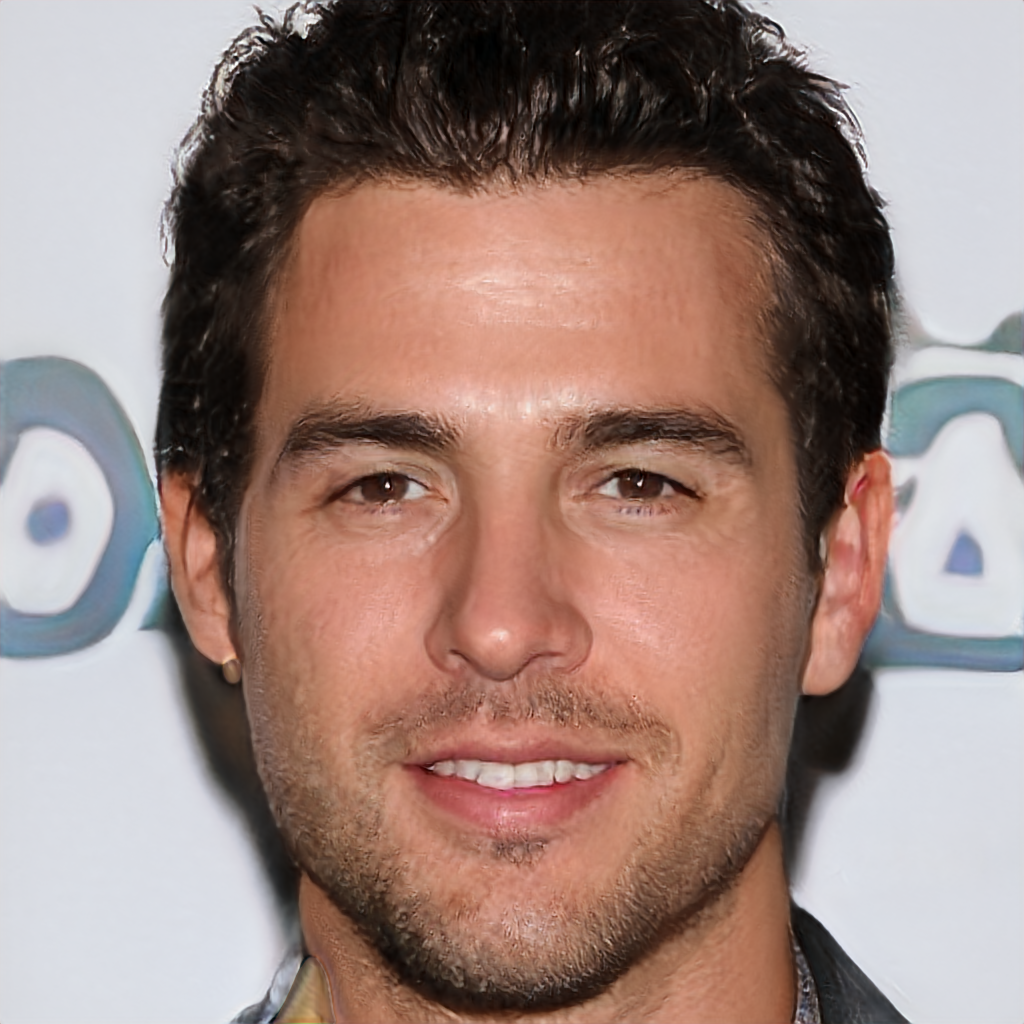}
		\includegraphics[width=0.075\textwidth]{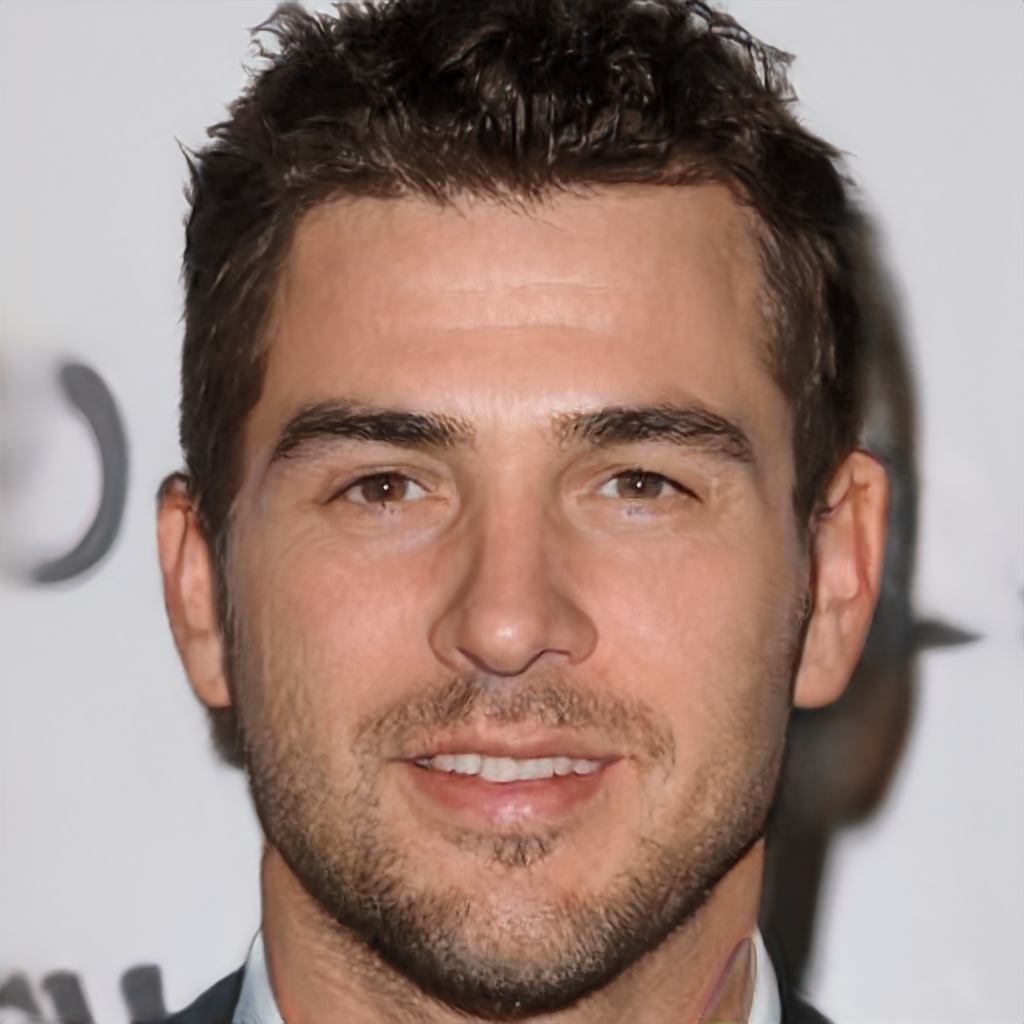}
		\includegraphics[width=0.075\textwidth]{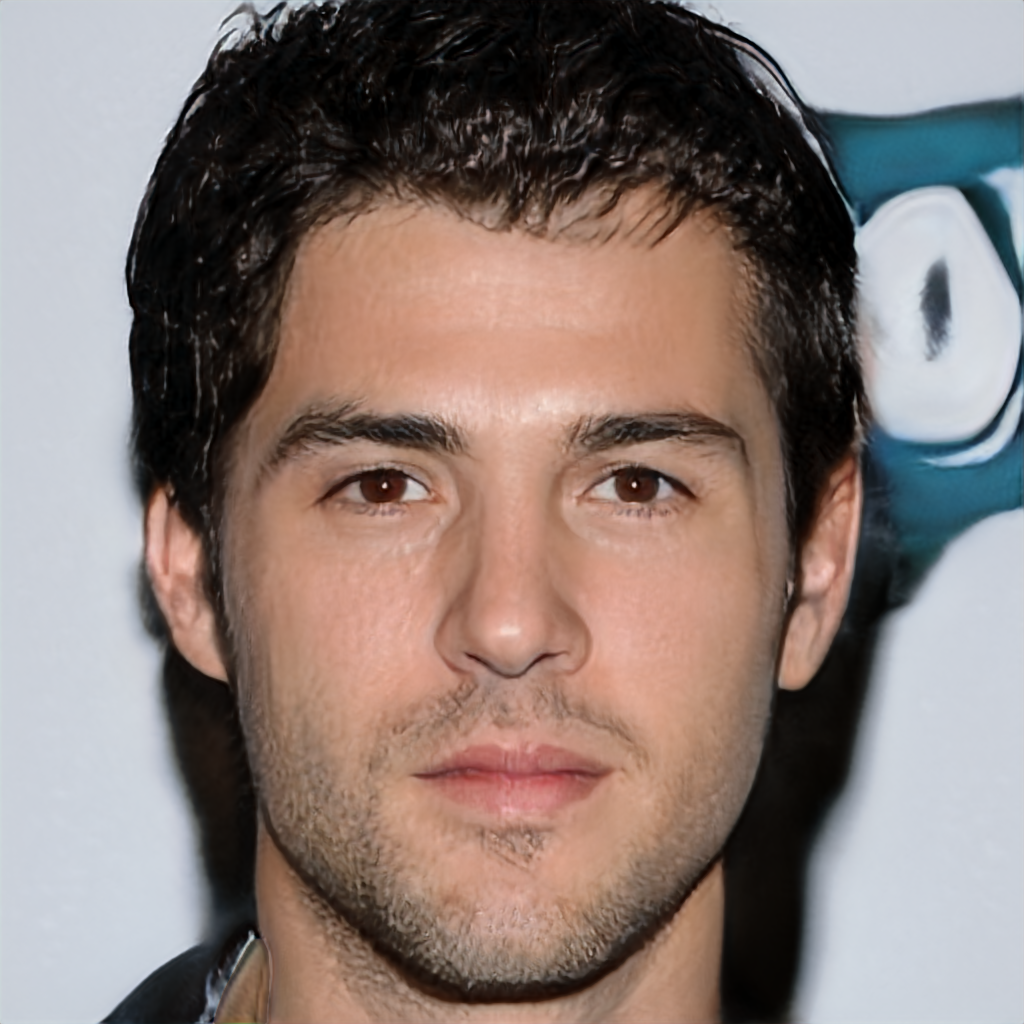}
		\includegraphics[width=0.075\textwidth]{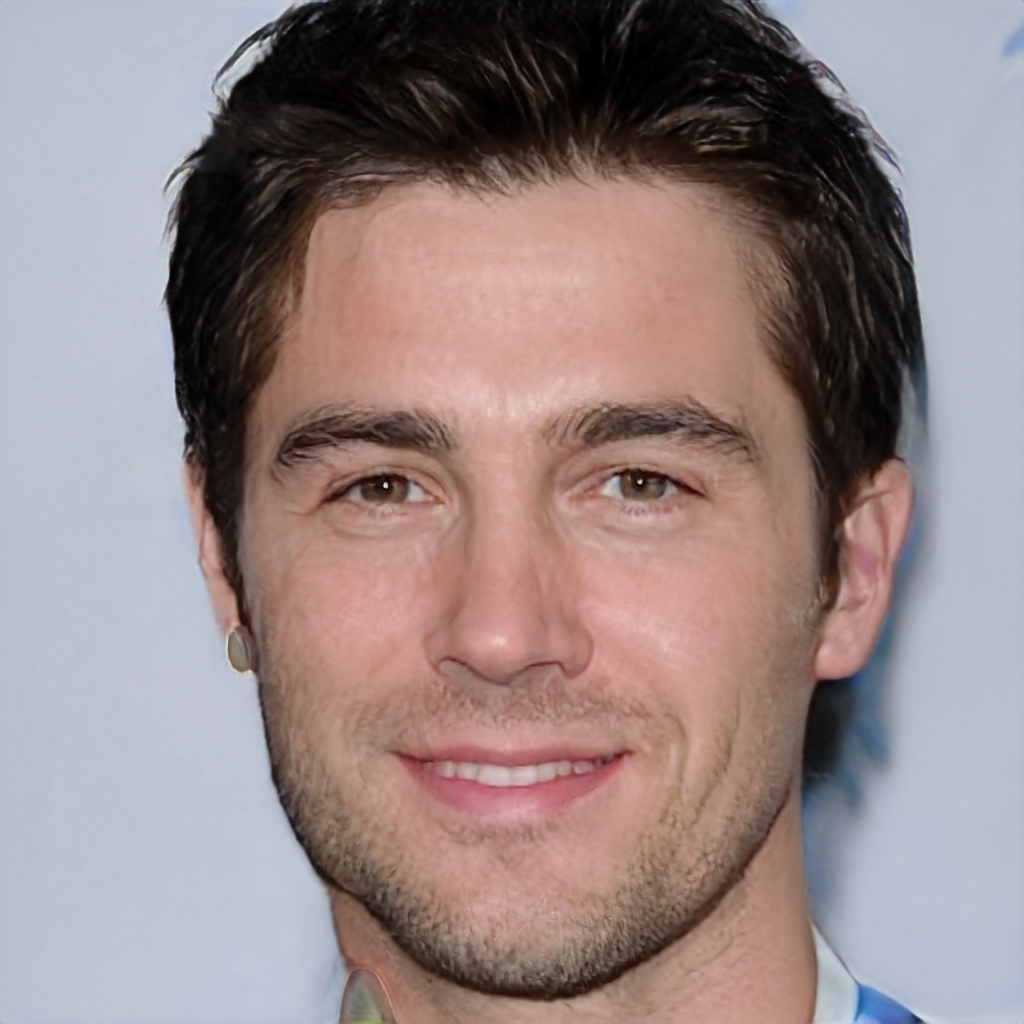}
		\includegraphics[width=0.075\textwidth]{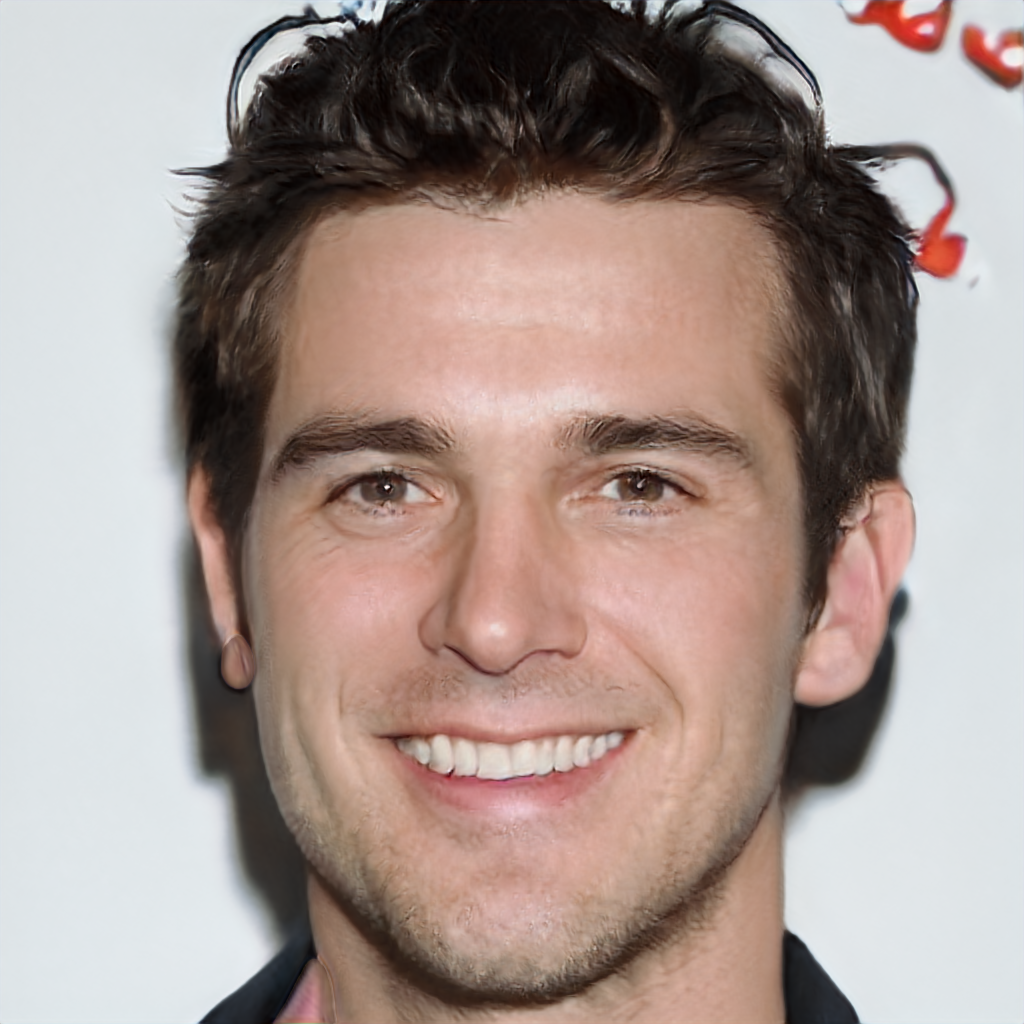} \\
		\includegraphics[width=0.075\textwidth]{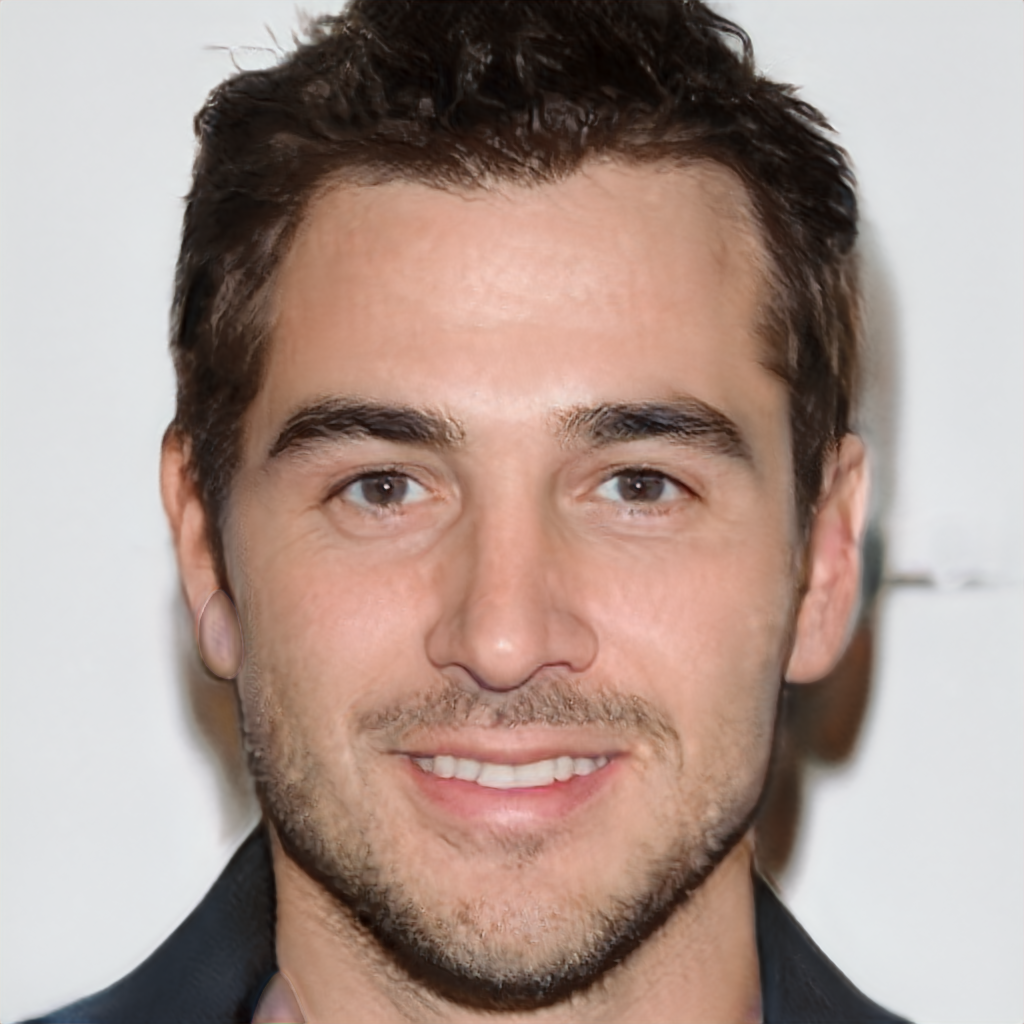}
		\includegraphics[width=0.075\textwidth]{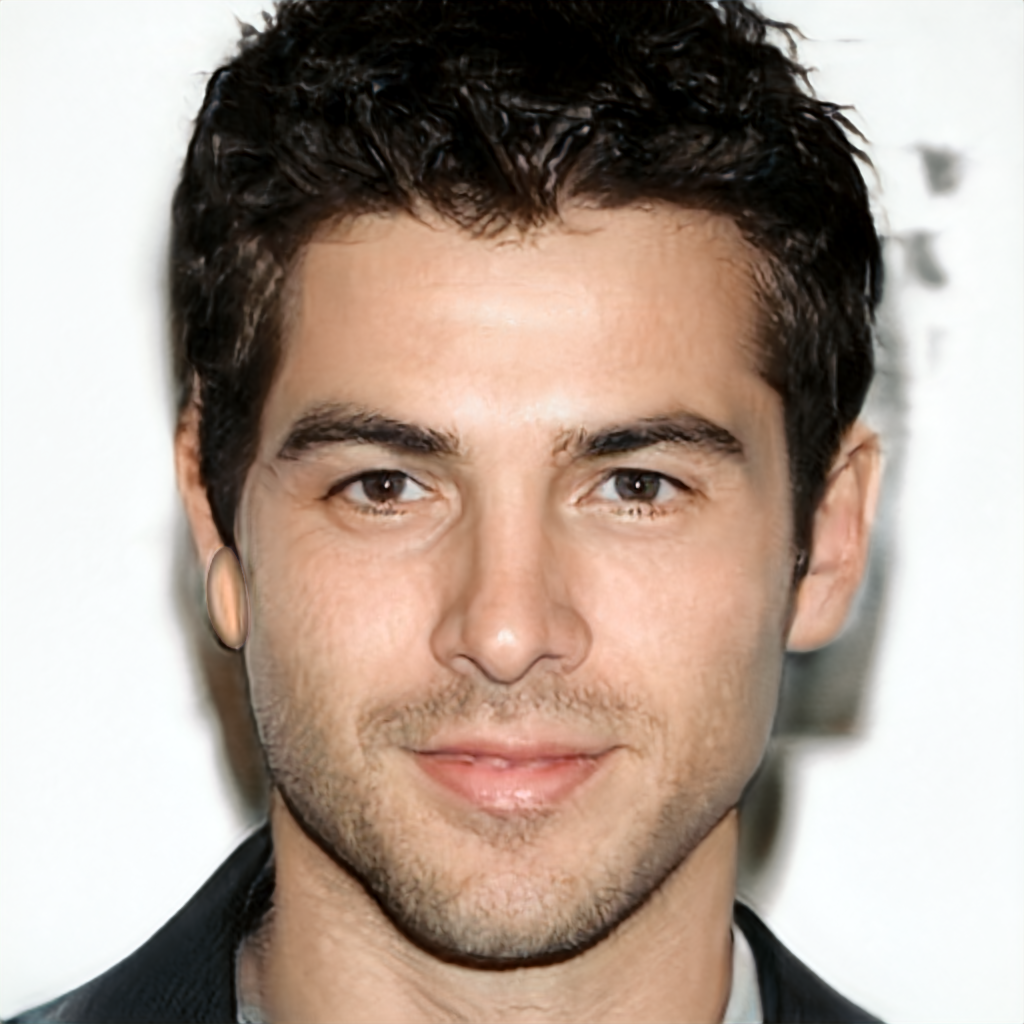}
		\includegraphics[width=0.075\textwidth]{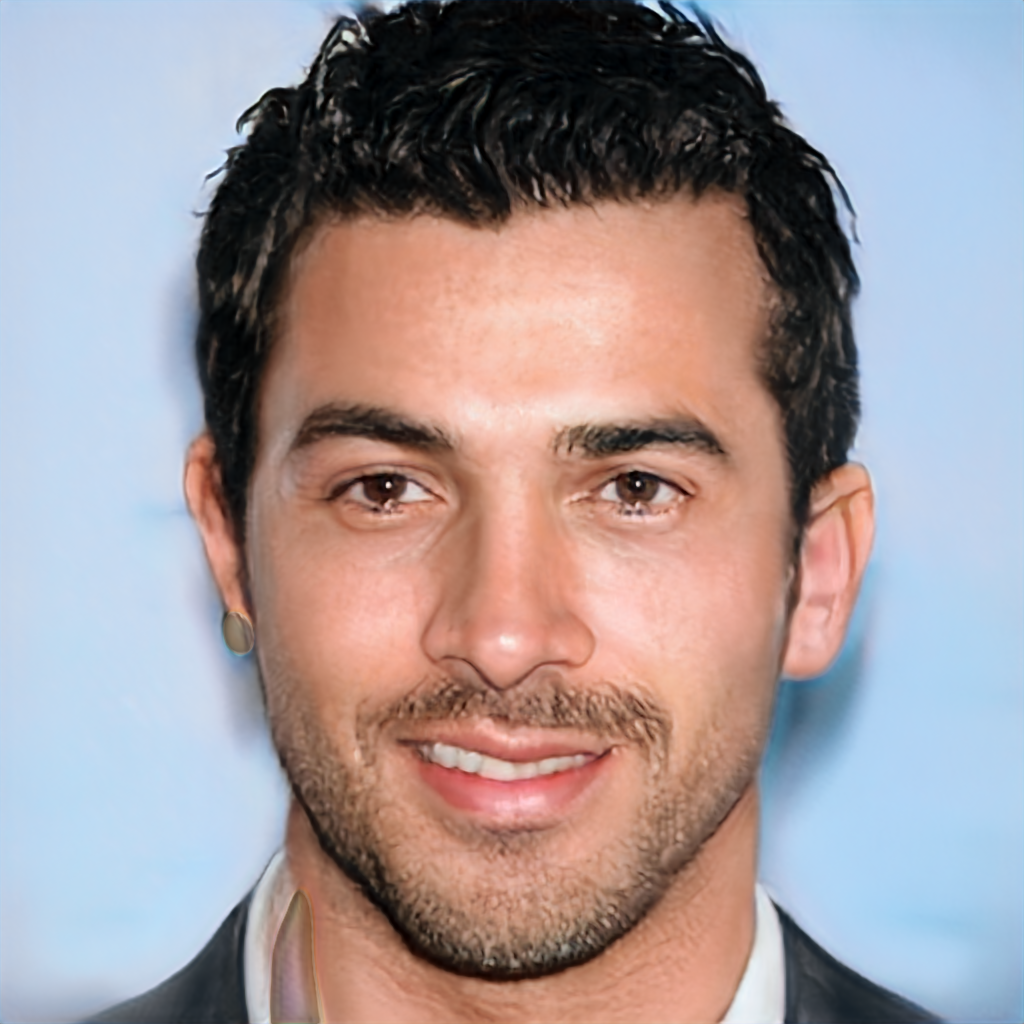}
		\includegraphics[width=0.075\textwidth]{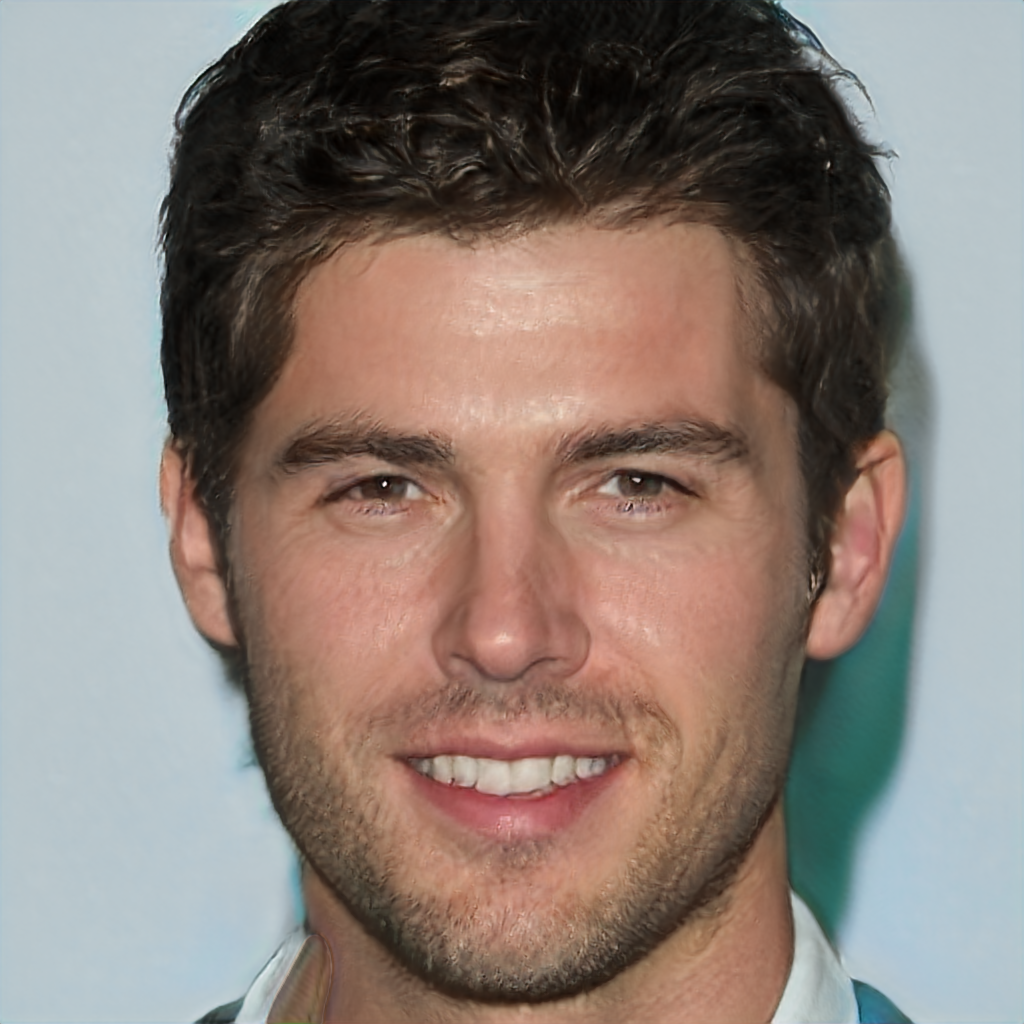}
		\includegraphics[width=0.075\textwidth]{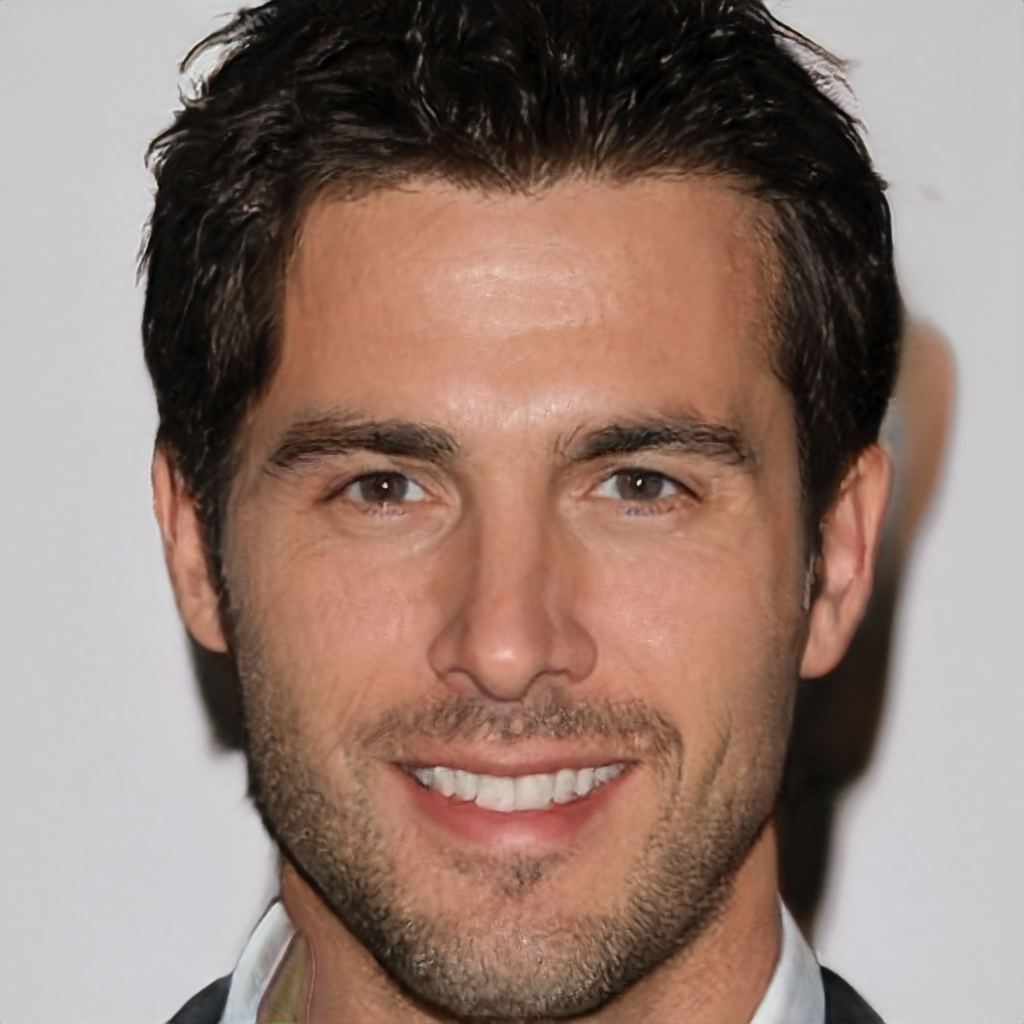}
		\includegraphics[width=0.075\textwidth]{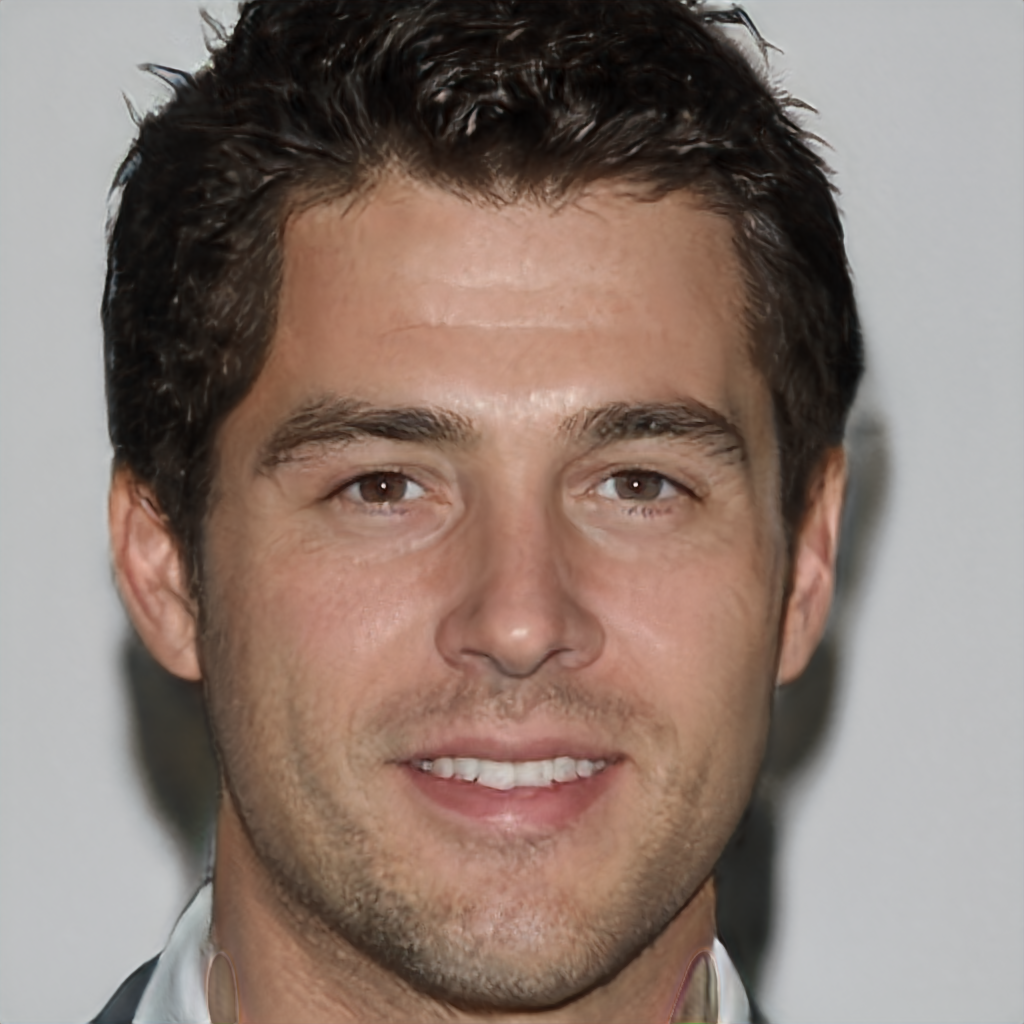}
		\includegraphics[width=0.075\textwidth]{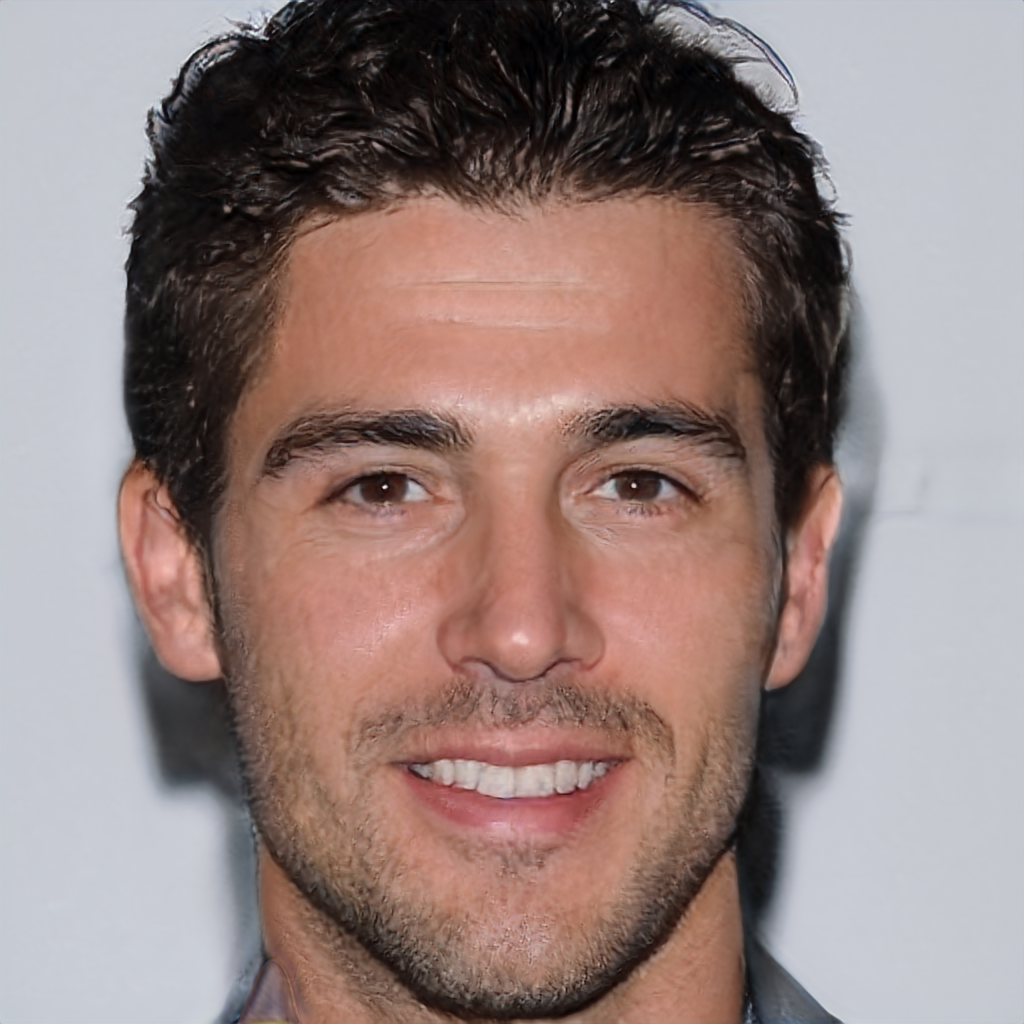}
		\includegraphics[width=0.075\textwidth]{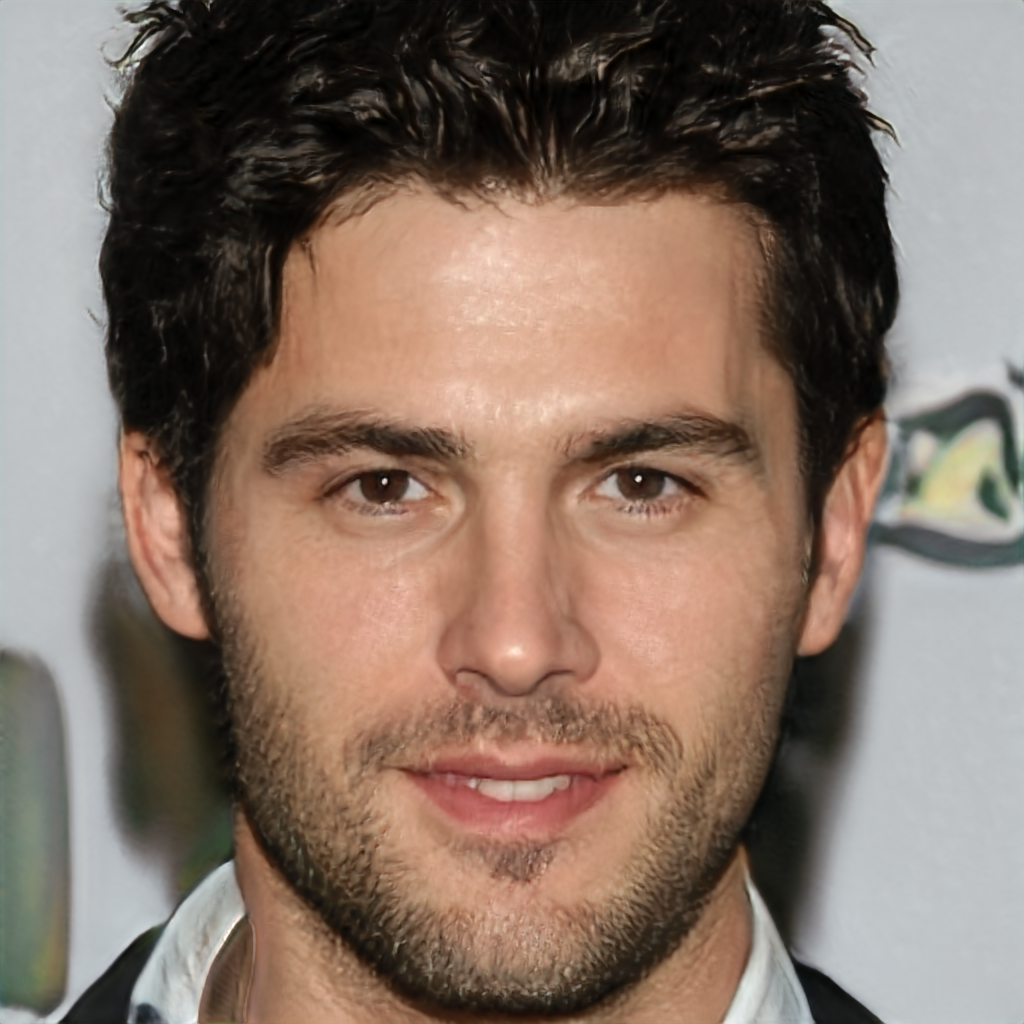}
		\includegraphics[width=0.075\textwidth]{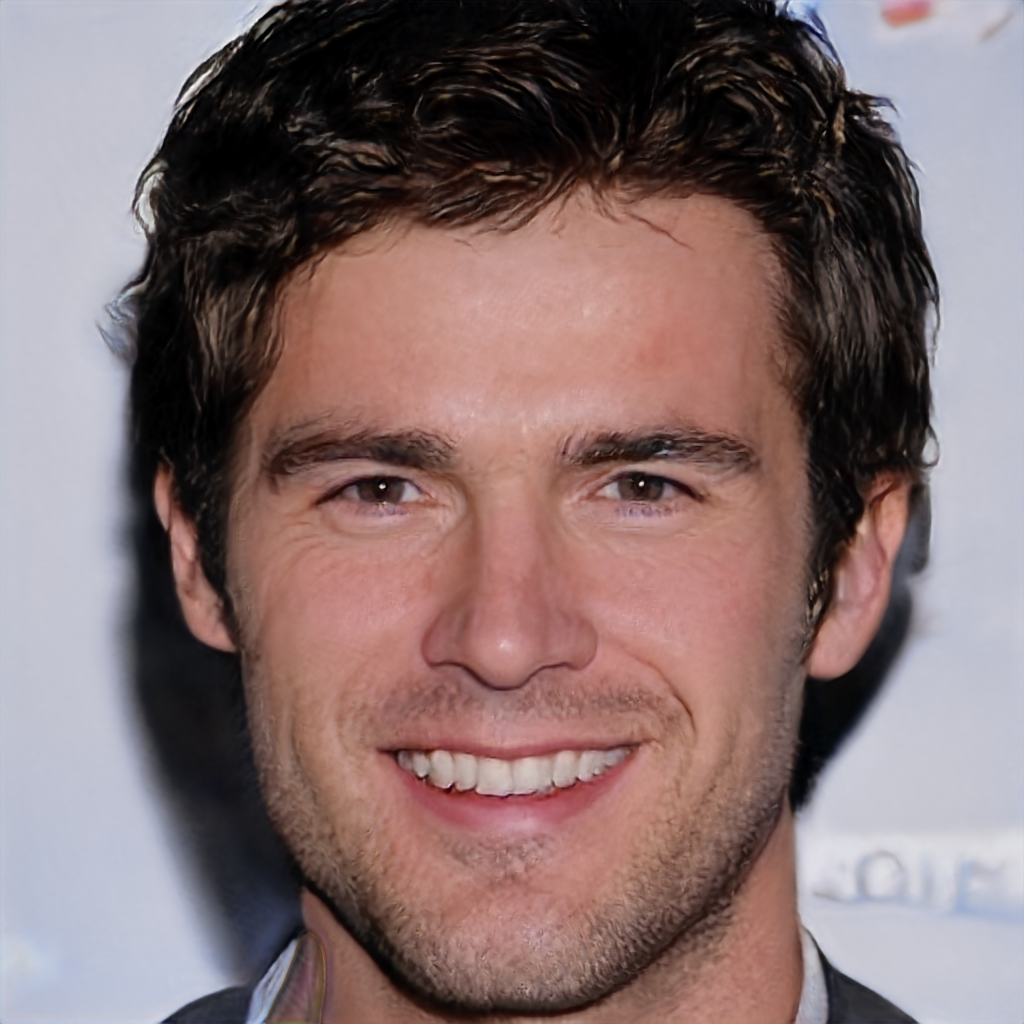}
		\includegraphics[width=0.075\textwidth]{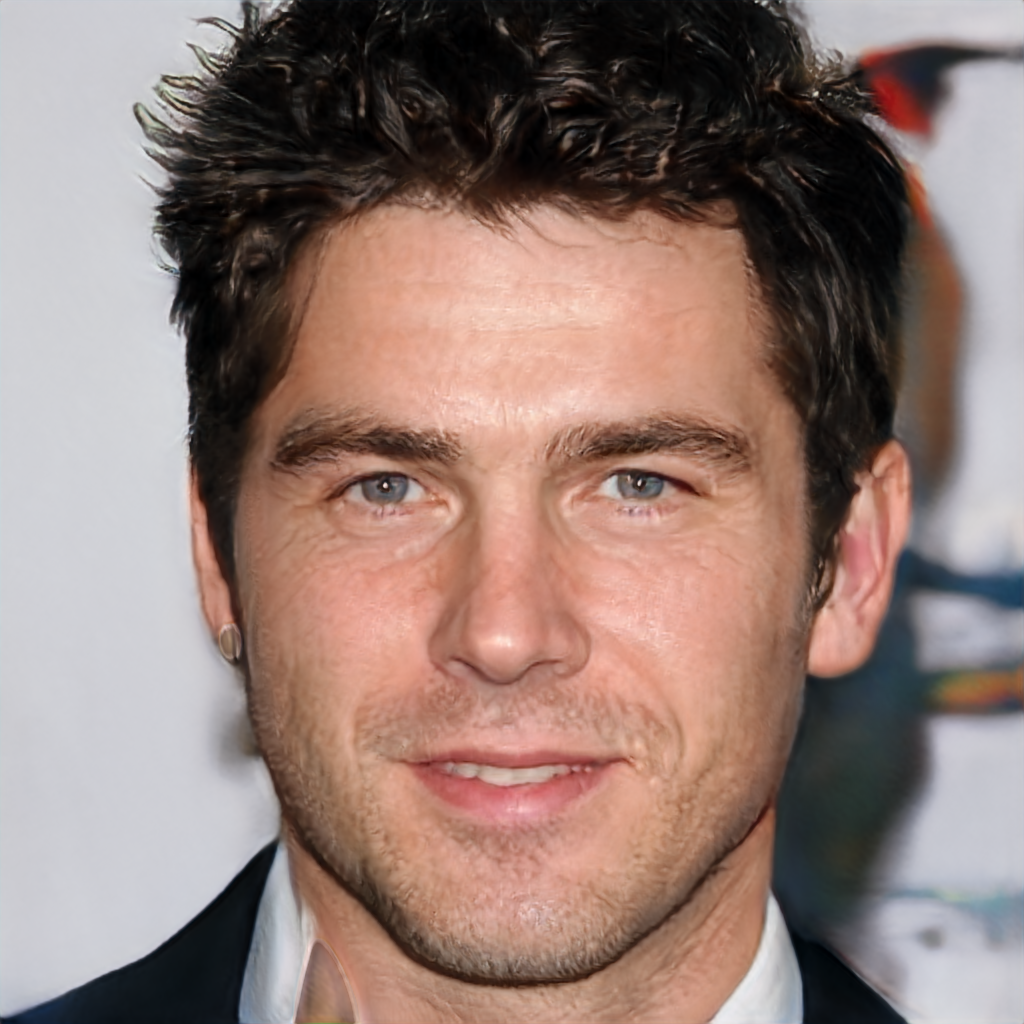}
		\includegraphics[width=0.075\textwidth]{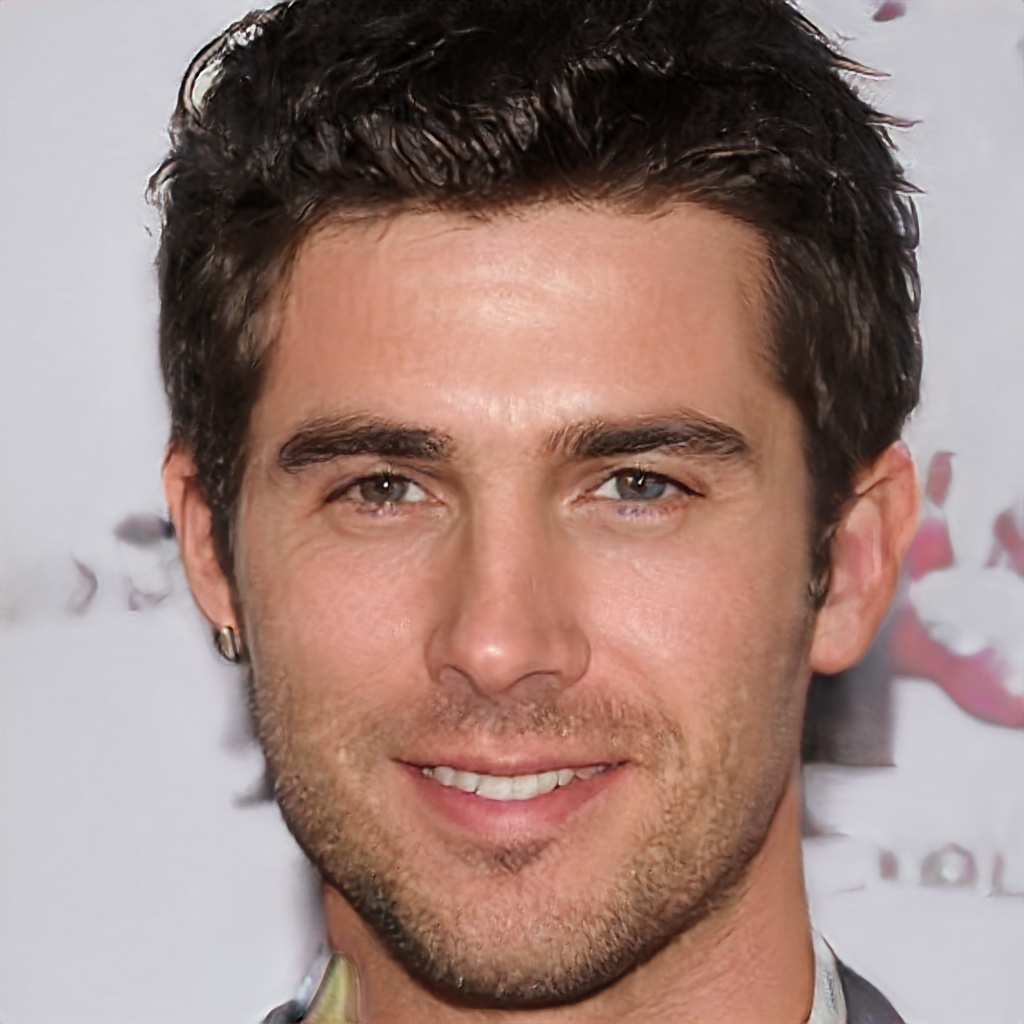}
		\includegraphics[width=0.075\textwidth]{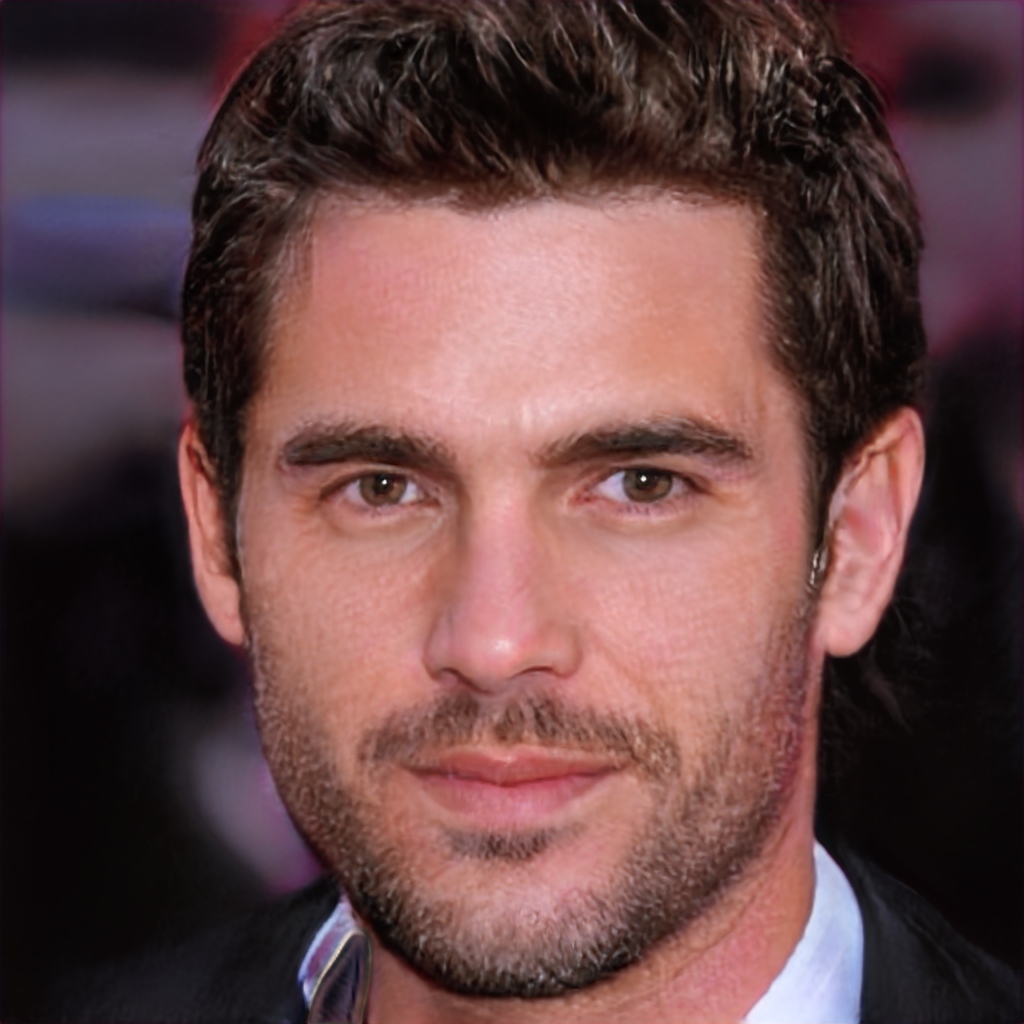}
		\label{fig:stb}
\end{tabular}
\vspace{-1.0em}
\caption{\small Using the $10$ million face images sampled from a StyleGAN trained on CelebAHQ-$1024$, we visualize the top $24$ modes with largest number of neighbors within $0.25$ distance described by $F_{id}$ and found that they looks very similar. Calibrating the \textit{worst-case} dense mode, which is the first mode shown here, implicitly takes the top $24$ modes dense modes into account.}
\label{fig:visualization}
\end{figure}